\documentclass[myepj-spec,pdftex, iicol]{mySvjour}
\usepackage{graphicx}
\usepackage[pdfpagemode=UseNone]{hyperref}
\usepackage{color}
\usepackage{xspace}
\usepackage[square,sort&compress,numbers]{natbib}   
\usepackage{verbatim}
\usepackage{amsmath,amssymb,amsfonts} % Typical maths resource packages
\usepackage{tabularx}
\usepackage{multicol}
\usepackage{footnote}
\usepackage{listings}
\usepackage[gen]{eurosym}
\usepackage[switch, modulo]{lineno}
\usepackage{longtable}
\usepackage[utf8]{inputenc} % allow utf-8 input
\usepackage[T1]{fontenc}    % use 8-bit T1 fonts
\usepackage{hyperref}       % hyperlinks
\usepackage{url}            % simple URL typesetting
\usepackage{booktabs}       % professional-quality tables
\usepackage{amsfonts}       % blackboard math symbols
\usepackage{nicefrac}       % compact symbols for 1/2, etc.
\usepackage{microtype}      % microtypography
\usepackage{xcolor}         % colors
\usepackage{graphicx}
\usepackage{float}
\usepackage{caption}
\usepackage{subcaption}
\usepackage{amsmath}
\usepackage{minted}
\usepackage{csquotes}
\setcitestyle{square}
\usepackage{svg}
\usepackage[title]{appendix}%

\usepackage{lmodern,bm}                
\usepackage[T1]{sansmath} 
\SetMathAlphabet{\mathsfbf}{sans}{\sansmathencoding}{\sfdefault}{bx}{sl}
\usepackage{etoolbox}
\AtBeginEnvironment{sansmath}{}{}{}

\usepackage{lipsum}

\definecolor{darkblue1}{rgb}{0,0,.2}
\definecolor{darkblue}{rgb}{0,0,.2}
\definecolor{darkred}{rgb}{0.5,0,0}
\pagecolor{white} % Background color
\color{black}     % Text color
\hypersetup{breaklinks=true, 
	colorlinks=true, 
	linkcolor=darkblue1, 
	menucolor=darkblue1, 
	urlcolor=darkblue1,
	citecolor=darkblue1,
	pdftitle={},
	pdfauthor={},
	pdfsubject={},
	pdfkeywords={},
	pdfproducer={}
}
\usepackage{siunitx}

\bibstyle{plain}
\begin{document}

			\begin{flushright}
				\normalsize
				%      \today
			\end{flushright}
			
			\vspace{-2cm}
			
			\title{\Large\boldmath Learning Minimal-Deviation Corrections for Multi-Dimensional Mismodelling in HEP Simulations}
\author{Matthias L. Schott$^1$\footnote{corresponding author: mschott@uni-bonn.de}, Lucie Flek$^2$}
\institute{$^1$ Institute of Physics, University of Bonn, Germany, $^2$ Bonn-Aachen International Center for Information Technology (b-it), University of Bonn, Germany}
			
\abstract{Accurate Monte Carlo (MC) modelling in high-energy physics is challenging, particularly in complex scenarios where simulations fail to reproduce observed data. In practice, experimental information is often limited to one-dimensional (1D) distributions, while mismodelling arises in a multidimensional feature space. This restricts traditional correction methods, as one-dimensional reweighting ignores correlations and fully multidimensional approaches require large target datasets. We propose a neural network–based method that operates under these constraints by learning a transformation of simulated events that reproduces the available 1D target distributions while remaining close to the original simulation. This minimal-deviation principle preserves the global correlation structure of the baseline model while enabling targeted corrections of mismodelled features. Using controlled studies with simulated pseudo-data, we show that the method improves agreement with target distributions and maintains a consistent multidimensional structure. The approach is designed for complex, high-dimensional analyses where traditional techniques are insufficient, providing a scalable way to enhance MC modelling under limited information.
}

\maketitle

\tableofcontents

\section{Introduction}

Monte Carlo (MC) event generators are a cornerstone of modern high-energy physics (HEP) analyses, providing detailed predictions for complex final states through a combination of perturbative calculations, parton showers, hadronization models, and phenomenological descriptions of soft QCD effects \cite{Sjostrand:2007gs}. In practical analyses, these simulations are typically combined with fast or full detector simulation in order to obtain realistic event samples for measurements and searches. Despite their central role, however, MC predictions are never perfect. Residual mismodelling between simulation and data is ubiquitous and is usually mitigated through data-driven corrections or tuning procedures \cite{Andreassen:2019nnm}.

A key practical limitation is that experimental information is often available only in the form of one-dimensional (1D) distributions for selected observables. While the underlying event structure is intrinsically multidimensional, precise multidimensional measurements are much harder to obtain because of detector effects, limited statistics, acceptance losses, and the complexity of unfolding in high dimensions \cite{Andreassen:2019cjw}. As a consequence, many corrections used in practice are derived from marginal distributions, even though the true differences between simulation and data generally involve non-trivial correlations across the full feature space.

Traditional reweighting techniques operate precisely in this regime. Event weights are adjusted such that simulated and observed distributions agree for one or a few chosen observables. This strategy is simple, robust, and widely used, but it leaves the event kinematics themselves unchanged and therefore cannot consistently repair mismodelling in correlated feature spaces \cite{Rogozhnikov:2016bdp}. More modern machine-learning-based reweighting methods can in principle exploit the full phase space and have shown impressive performance in collider applications. However, these methods usually assume that representative source and target samples are available in the full multidimensional space, which is often unrealistic when only low-dimensional measurements can be extracted reliably from data \cite{Cranmer:2015bka, Nachman:2021opi}.

In this work, we consider a different perspective motivated by the observation that modern MC generators already provide a reasonable approximation of the global event structure, including many physically meaningful correlations. In many applications—particularly exploratory measurements, searches, or studies of complex environments such as the underlying event—the main challenge is therefore not to construct a completely new generative model, but rather to correct specific deficiencies of an otherwise adequate baseline simulation \cite{Daumann:2024kfd}.

We formulate this task as a constrained transformation of simulated events under limited information. Instead of learning an arbitrary mapping to data, we seek a transformation that satisfies two complementary principles: first, agreement with observed data at the level of the available information, meaning that the transformed simulation reproduces the measured 1D distributions of relevant observables; second, minimal deviation from the original simulation, such that the learned transformation remains as close as possible to the identity map. This second requirement acts as a physically motivated regularization. Among the many possible transformations that reproduce a given set of marginals, we favor those that introduce the smallest modifications to the original events.

This viewpoint is conceptually related to optimal transport, where one seeks maps between probability distributions that minimize a transport cost \cite{peyre2020computationaloptimaltransport}. It is also connected to recent flow-based approaches that learn explicit transformations between simulated and target samples in HEP \cite{Daumann:2024kfd}. Our setting is more restricted, however: unlike most transport- or flow-based methods, we do not assume that the full multidimensional target distribution is available for training. Instead, we operate in an intermediate regime in which only partial information about the target distribution is known, while preserving the pre-existing structure of the baseline simulation is elevated to an explicit design principle.

This distinguishes the proposed method both from standard reweighting approaches, which alter event weights but not event kinematics, and from fully generative approaches, which aim to learn the entire multidimensional target distribution \cite{Papamakarios:2019fms, Cranmer:2019eaq}. The goal here is not to replace the baseline generator, but to refine it in a controlled way using the limited experimental information that is typically available.

It is important to emphasize that this method is not intended for ultra-high-precision measurements where systematic effects must be controlled at the per-mille level. Rather, it is designed for complex, high-dimensional scenarios in which traditional correction strategies become insufficient or impractical. In such settings, the proposed framework offers a scalable way to improve agreement between simulation and data while maintaining a physically plausible event structure. Moreover, because the size of the learned transformation is explicitly controlled, the method also provides a natural handle for assessing modelling uncertainties associated with imperfect simulations.

\section{Related Work}

Correcting mismodelling between simulated and observed distributions has long been an important problem in HEP and in other data-intensive scientific domains. Existing approaches range from simple histogram-based corrections to modern machine-learning methods that estimate density ratios or learn explicit transformations between probability distributions \cite{Andreassen:2019nnm, peyre2020computationaloptimaltransport, Papamakarios:2019fms}.

The most widely used correction strategy in collider physics remains one-dimensional reweighting. In this approach, event weights are derived from the ratio of target and simulated distributions for a chosen observable, such as a particle transverse momentum or an event-level quantity like pileup multiplicity. This procedure is straightforward and often robust in practice, but it only enforces agreement for the selected marginal distribution and does not modify correlations with other observables \cite{Rogozhnikov:2016bdp}. As a result, significant mismodelling can remain in multidimensional feature spaces even after excellent closure is achieved at the 1D level.

A closely related one-dimensional transformation technique is quantile mapping, also referred to as histogram matching. Here, samples are transformed through their cumulative distribution functions such that the corrected distribution reproduces a desired target histogram. Quantile mapping has been used extensively in climate science for bias correction of simulated temperature and precipitation distributions \cite{cannon2015bias}. In one dimension, this construction is closely tied to the monotone optimal transport map and can be interpreted as a minimal-displacement transformation under suitable cost functions \cite{peyre2020computationaloptimaltransport}.

To move beyond purely marginal corrections, several multidimensional reweighting strategies have been developed. Early approaches based on multidimensional histograms or kernel density estimation become rapidly impractical as dimensionality increases. More recently, machine-learning models such as boosted decision trees and neural networks have been used to learn reweighting factors directly in high-dimensional spaces. In HEP, boosted-decision-tree reweighting and neural full-phase-space reweighting have demonstrated that rich multidimensional corrections can be learned efficiently when suitable target samples are available \cite{Nachman:2021opi, Cranmer:2015bka}. Related ideas also appear in iterative unfolding approaches such as OmniFold, which uses classifier-based reweighting to exploit the full phase space in detector corrections \cite{Andreassen:2019cjw}.

Classifier-based reweighting methods are particularly important in this context. Their conceptual basis is closely related to density-ratio estimation: a classifier trained to distinguish source from target events can be converted into an estimate of the corresponding likelihood or density ratio \cite{CMS:2024jdl}. This perspective has become central in simulation-based inference and in modern HEP reweighting methods. Extensions such as neural conditional reweighting further generalize this idea to settings where one wishes to condition on auxiliary features instead of marginalizing over them \cite{lopezpaz2018revisitingclassifiertwosampletests}. These methods are highly flexible, but in their standard form they still rely on representative multidimensional target information for training.

That requirement becomes a serious limitation in realistic experimental situations. As the dimensionality of the feature space grows, the statistics needed to constrain multidimensional corrections increase rapidly, and the learned weights can become unstable in sparsely populated regions. This problem is especially acute when one attempts to learn corrections directly from data, where only selected projected distributions may be known with sufficient precision. Our work is motivated precisely by this gap between the information that is typically available in analyses and the information required by full multidimensional reweighting.

The broader mathematical problem of transforming one distribution into another while minimizing a cost function is studied in the framework of optimal transport (OT) \cite{peyre2020computationaloptimaltransport}. OT has become increasingly influential in machine learning because it provides a principled way to compare and align probability distributions, including in settings such as domain adaptation, generative modeling, and dataset alignment. In principle, OT provides exactly the kind of minimal-movement notion that is attractive for correcting simulated events. In practice, however, OT-based methods generally assume access to the full source and target distributions, which is not the case in many HEP applications.

Neural-network-based distribution transport methods, including normalizing flows, provide another important point of contact \cite{Papamakarios:2019fms}. Flows learn explicit invertible transformations between distributions and have become a standard tool for high-dimensional probabilistic modeling \cite{Cranmer:2019eaq}. Very recently, such ideas have also been used directly for simulation correction in HEP, where a normalizing flow is trained to transform a simulated distribution into a target one while preserving non-trivial correlations \cite{Daumann:2024kfd}. These approaches are highly expressive, but again they typically rely on full multidimensional target samples and therefore address a different information regime from the one considered here.

Outside HEP, a related challenge appears in multivariate bias correction, especially in climate science. There, one often seeks to adjust marginal distributions while preserving or restoring dependence structures between variables. Multivariate extensions of quantile mapping and stochastic bias-correction methods based on optimal transport have been developed precisely for this purpose \cite{maraun2015bias}. These methods are conceptually close to our problem because they recognize that correcting marginals alone is insufficient when correlations matter. The main difference in our case is that the baseline simulation already contains substantial physically motivated structure, which we explicitly aim to preserve by penalizing large deviations from the identity mapping.

\section{Methodology}
\begin{figure}
    \centering
    \includegraphics[width=0.49\textwidth]{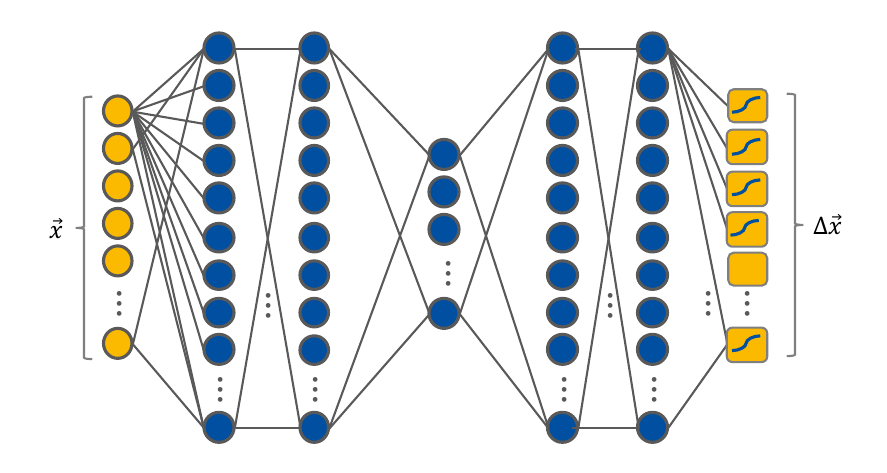}
    \includegraphics[width=0.49\textwidth]{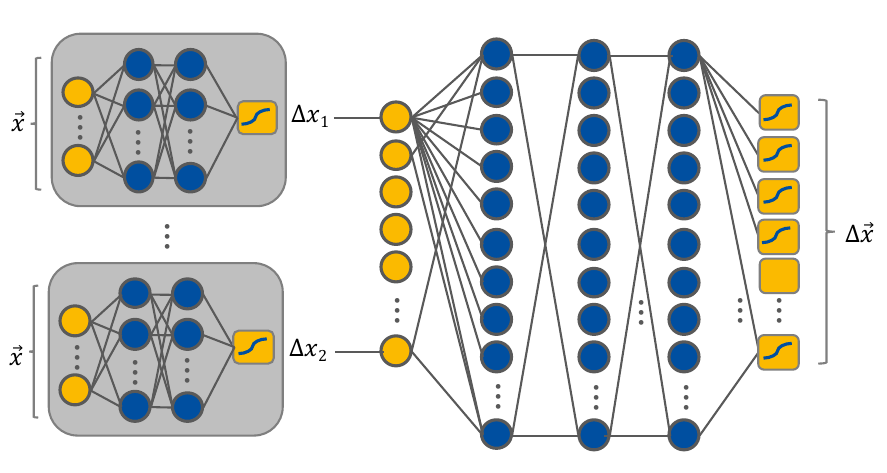}
    \caption{Schematic illustration of the network architectures: the Global Residual Transformation (left), which learns a single unified mapping, and the Two-Step Residual Transformation (right), where the transformation is factorized into two sequential residual updates.}
    \label{fig:NNArch}
\end{figure}

We consider the problem of transforming an input sample $x \in \mathbb{R}^d$ into a modified sample $x'$ such that its statistical properties better agree with a given target distribution, while preserving the essential structure of the original data. This setting is motivated by typical scenarios in high-energy physics where an existing model captures the dominant features and correlations, but requires controlled corrections to better describe observed data.
Across all approaches, we adopt a \emph{residual transformation framework}, in which the modified sample is obtained as
\begin{equation}
x' = x + \Delta(x),
\end{equation}
where $\Delta(x)$ denotes a learned correction that is explicitly constrained to remain small and structured.
Two complementary strategies are considered: (i) a \emph{global residual model} acting on all features simultaneously (Figure \ref{fig:NNArch}), and (ii) a \emph{two-step approach} combining feature-wise corrections with a subsequent global refinement (Figure \ref{fig:NNArch}).

\subsection{Global Residual Transformation}
In the global approach, a single neural network $\Delta_\theta : \mathbb{R}^d \to \mathbb{R}^d$ predicts corrections for all features jointly. To ensure stability and locality, the correction is bounded component-wise,
\begin{equation}
\Delta_{\theta,j}(x) = \alpha_j \tanh\big(f_{\theta,j}(x)\big),
\end{equation}
where $f_\theta$ is a multi-layer perceptron and $\alpha_j$ defines the maximal allowed modification per feature. Optionally, a binary mask $m \in {0,1}^d$ restricts the transformation to selected features. The model is trained using a composite loss function,
\begin{equation}
\mathcal{L} =
\lambda_{\mathrm{hist}} \mathcal{L}_{\mathrm{hist}} +
\lambda_{\mathrm{der}} \mathcal{L}_{\mathrm{der}} +
\lambda_{\mathrm{move}} \mathcal{L}_{\mathrm{move}} +
\lambda_{\mathrm{corr}} \mathcal{L}_{\mathrm{corr}},
\end{equation}
which balances four objectives:

A differentiable histogram loss, $\mathcal{L}_{\mathrm{hist}}$ is used to align the distributions of individual input features between transformed and target samples. In the context of high-energy physics, these features typically correspond to kinematic properties of reconstructed objects, such as the transverse momentum ($p_T$) of muons, the energy or mass of jets, angular variables like pseudorapidity ($\eta$) or azimuthal angle ($\phi$), or missing transverse energy. These quantities are often well measured and their one-dimensional distributions are known with relatively high precision from data.
To compare such distributions during training, we employ a \emph{differentiable histogramming} procedure. Instead of assigning each event to a discrete bin (as in a standard histogram), each entry contributes smoothly to neighboring bins via a sigmoid-based soft assignment. \footnote{For a bin with lower and upper edges $b_i$ and $b_{i+1}$, respectively, the soft assignment of an entry $x$ is defined as
$w_i(x) = \sigma\!\left((x-b_i)/\tau_i\right) - \sigma\!\left((x-b_{i+1})/\tau_i\right)$,
where $\sigma(z)=(1+e^{-z})^{-1}$ is the sigmoid function and $\tau_i$ controls the smoothing scale. This difference of sigmoids provides a smooth approximation to the indicator function of a histogram bin, allowing entries near bin boundaries to contribute fractionally to neighboring bins. In the limit $\tau_i\to 0$, the assignment approaches conventional hard binning, while finite $\tau_i$ ensures differentiability with respect to $x$.} This results in a continuous approximation of the histogram, allowing gradients to propagate back through the binning operation.
This is essential, since a naïve $\chi^2$ comparison of binned histograms, is not differentiable with respect to the input features: small changes in $x$ do not affect the bin counts unless a bin boundary is crossed, leading to vanishing gradients almost everywhere. As a consequence, such a loss cannot effectively guide the training of a neural network. The differentiable histogram loss overcomes this limitation by providing smooth, non-zero gradients, enabling stable and efficient optimization.

In addition to individual features, we impose constraints on a set of derived observables ${\mathcal{O}_\ell(x)}$, via the term $\mathcal{L}_{\mathrm{der}}$, which encode higher-level, physically meaningful properties of the event. These observables are typically constructed from multiple input features and probe correlations between them. Examples include invariant masses of particle systems (e.g.\ dilepton or dijet invariant masses), angular separations such as $\Delta R$ between objects, transverse mass variables, or global event shape observables.
For instance, the invariant mass of two muons depends jointly on their momenta and angular separation. Even if the individual $p_T$ distributions are well modeled, mismodelling in their correlations can lead to incorrect invariant mass distributions. By explicitly matching the distributions of such derived quantities, the model is encouraged to correct not only marginal feature distributions but also their joint structure.
As for the feature-wise case, differentiable histogram losses are used to compare predicted and target distributions of these observables. This ensures that gradients can flow through the full computational chain—from the observable definition back to the individual input features—allowing the network to learn coordinated adjustments across multiple variables. This is particularly important in scenarios where correlations are only approximately modeled in the input and must be refined to achieve agreement with data.

The transformation is constrained to remain close to the original input, reflecting the assumption that the baseline already provides a reasonable description. This is enforced via a movement penalty, $\mathcal{L}_{\mathrm{move}} \propto |x' - x|^2$, which regularizes the residual correction and suppresses large, unphysical shifts. Feature-wise weights can be introduced to encode different levels of trust in individual observables (e.g.\ tighter constraints for well-measured quantities such as muon momenta). Invalid or padded entries can be excluded to avoid introducing artificial structure.

To maintain the multivariate structure, deviations in the empirical correlation matrix are penalized, $\mathcal{L}_{\mathrm{corr}} = | \rho(x') - \rho(x) |_F^2$. Here, $\rho(\cdot)$ denotes the matrix of pairwise (one-dimensional) Pearson correlation coefficients between all features, which is computed for both the transformed and reference samples during training. This ensures that linear dependencies between features are preserved, preventing distortions of higher-level behavior. The term reflects the prior that the dominant correlations are already well modeled, and that only minimal, coordinated corrections are required.

\subsection{Two-Step Residual Transformation}
As an alternative, we decompose the transformation into two stages, separating feature-wise corrections from global consistency enforcement. In the first stage, independent residual models are trained for each feature. For feature $j$, the correction is defined as
\begin{equation}
\label{eqn:34alt}
\Delta_j(x) = \alpha_j \tanh\big(f_{\theta_j}(x_{I_j})\big),
\end{equation}
where $x_{I_j}$ denotes a selected subset of input features providing contextual information.
The transformed feature is
\begin{equation}
x'_j = x_j + \Delta_j(x),
\end{equation}
with the correction optionally suppressed for vanishing inputs to respect sparsity patterns.
Each model is trained independently using
\begin{equation}
\mathcal{L}j =
\mathcal{L}_{\mathrm{hist},j}
+
\lambda_{\mathrm{move}} , \mathbb{E}\big[\Delta_j(x)^2\big],
\end{equation}
where $\mathcal{L}_{\mathrm{hist},j}$ is a differentiable histogram loss matching the distribution of $x'_j$ to the target. The second term,
\begin{equation}
\lambda_{\mathrm{move}}\,\mathbb{E}\!\left[\Delta_j(x)^2\right],
\end{equation}
acts as a movement penalty that regularizes the learned transformation. Here, $\mathbb{E}[\Delta_j(x)^2]$ denotes the mean squared magnitude of the residual correction over the training sample, while $\lambda_{\mathrm{move}}$ controls the strength of this regularization. Minimizing this term favors solutions that reproduce the target distribution with the smallest possible modifications to the original feature values. This stage focuses on accurately reproducing one-dimensional feature distributions with minimal modifications, while keeping the models low-dimensional and interpretable. Correlations are only captured implicitly through the choice of input subsets $I_j$.

In the second stage, a global residual network is applied to the output of the first stage,
\begin{equation}
x'' = x' + \Delta_\phi(x'),
\end{equation}
with the same bounded residual structure as in the global approach.
In contrast to the first stage, the training objective emphasizes higher-level consistency. The loss is defined as
\begin{equation}
\mathcal{L}{\mathrm{total}} =
\mathcal{L}{\mathrm{der}}
+
\lambda_{\mathrm{move}} \mathcal{L}_{\mathrm{move}}
+
\lambda_{\mathrm{corr}} \mathcal{L}_{\mathrm{corr}},
\end{equation}
where $\mathcal{L}_{\mathrm{der}}$ enforces agreement of derived observables via differentiable histogramming,
$\mathcal{L}_{\mathrm{move}}$ penalizes the size of the residual correction relative to allowed feature-wise bounds,
$\mathcal{L}_{\mathrm{corr}}$ aligns the correlation matrix with that of the target sample.
This refinement stage enables coordinated adjustments across all features, correcting residual mismodelling in correlations and derived quantities that cannot be captured by independent feature-wise transformations.
\subsection{Summary of the Approaches}
Both strategies share the same underlying principle of \emph{bounded residual corrections}. The global approach performs a single, fully coupled transformation that directly enforces all constraints simultaneously. In contrast, the two-step approach first solves a set of simpler feature-wise problems and subsequently restores global consistency through a dedicated refinement network.
The latter is particularly advantageous in complex scenarios where one-dimensional distributions are well understood, while correlations and higher-level structures are only approximately modeled.

\section{Experimental Setups}

To evaluate the proposed approach, we construct controlled benchmark scenarios in which a baseline Monte Carlo (MC) sample is transformed to match a target MC sample of the same physical process, but generated with different model assumptions. This setup allows for a quantitative assessment, since the target distribution is fully known.
In all cases, events are generated using \textsc{Pythia8} \cite{Sjostrand:2007gs} and passed through a fast detector simulation based on \textsc{Delphes} \cite{deFavereau:2013fsa}. The transformation is trained using only one-dimensional feature distributions, while performance is evaluated both at the level of these features and on higher-level observables. This allows us to explicitly test whether the method can recover multivariate structure beyond the information provided during training.
We consider two complementary physics scenarios with very different underlying modeling challenges:

As a first benchmark, we study $t\bar{t}$ production in proton--antiproton collisions at a center-of-mass energy of $\sqrt{s} = 1.96$ TeV (Tevatron-like conditions) as the baseline, and aim to transform it to $t\bar{t}$ production in proton--proton collisions at $\sqrt{s} = 13$ TeV (LHC conditions). This setup probes the ability of the method to learn non-trivial kinematic shifts induced by a large change in energy scale and parton luminosities.
The input features consist of reconstructed object kinematics, including missing transverse energy ($E_T^{Miss}$) and its azimuthal angle, as well as the kinematic properties of the two leading muons and the four leading jets (e.g.\ transverse momentum, pseudorapidity, and azimuthal angle). As derived observables, we consider composite quantities sensitive to correlations, such as the invariant mass of the muon pair, the invariant mass of the two leading jets, and the transverse momentum of multi-jet systems.

As a second benchmark, we consider a qualitatively different regime focusing on soft QCD physics and underlying event modeling. Here, the baseline sample is generated using the AZ tune \cite{ATLAS:2014alx}, while the target sample is based on the Monash tune \cite{Skands:2014pea} within \textsc{Pythia8}. Both samples correspond to soft QCD (minimum bias) processes in proton--proton collisions at $\sqrt{s} = 13$ TeV.
This scenario is particularly challenging, as differences between tunes are subtle and predominantly affect soft, diffuse activity rather than hard, well-defined objects. It therefore provides a stringent test of whether the method can learn small but structured corrections in a high-dimensional and weakly constrained environment.

A summary of the generated data sets and their configurations is provided in Table~\ref{tab:mc_samples}. In both setups, the comparison between transformed and target samples is performed at the level of input features, derived observables, and correlation structure, allowing for a comprehensive evaluation of the proposed methodology.

\begin{table}
\centering
\begin{tabular}{l | l | c | c | l }
\hline
Process & Generator & Order & $\sqrt{s}$ [TeV] & PDF / Tune \\
\hline

$pp \rightarrow t\bar{t} X \rightarrow l^+\nu l^- \bar\nu b \bar{b} + X$ 
& \textsc{Pythia8} & LO & 13 & CTEQ6.6 \\

$p\bar{p} \rightarrow t\bar{t} X \rightarrow l^+\nu l^- \bar \nu b \bar{b} + X$ 
& \textsc{Pythia8} & LO & 1.96 & CTEQ6.6 \\

\hline

$pp \rightarrow X$ (U.E.) 
& \textsc{Pythia8} & LO+Non.Pertb. & 13 & CTEQ6.6 / AZ-Tune \\

$pp \rightarrow X$ (U.E.) 
& \textsc{Pythia8} & LO+Non.Pertb. & 13 & CTEQ6.6 / Monash-Tune \\
\hline
\end{tabular}
\caption{Summary of simulated processes and their configurations. Each sample consists of 200k events.}
\label{tab:mc_samples}
\end{table}

\section{Results}

\subsection{Transfer from Top Quark Pair Production Processes from Tevatron to the LHC via a Global Residual Transformation}

In this study, we apply the global residual transformation approach to map $t\bar{t}$ events from Tevatron conditions ($\sqrt{s} = 1.96$ TeV, proton--antiproton collisions) to LHC conditions ($\sqrt{s} = 13$ TeV, proton--proton collisions).
The transformation is modeled using a fully connected neural network with four hidden layers and 256 neurons per layer. The network operates on 30 input features and predicts 30 residual corrections. As described in Section~X, the final output is obtained via a bounded residual update, ensuring that the transformation remains controlled. Input entries that are exactly zero (e.g.\ due to padding or missing objects) are explicitly protected and remain unchanged during the transformation.
The data set is split into 80\% training and 20\% validation samples. Training is performed using large batch sizes (greater than $5 \times 10^3$ events) to ensure stable estimates of the differentiable histogram losses. Early stopping is employed, terminating the training if no improvement in the validation loss is observed for five consecutive epochs.

\begin{figure}
    \centering
    \includegraphics[width=0.32\textwidth]{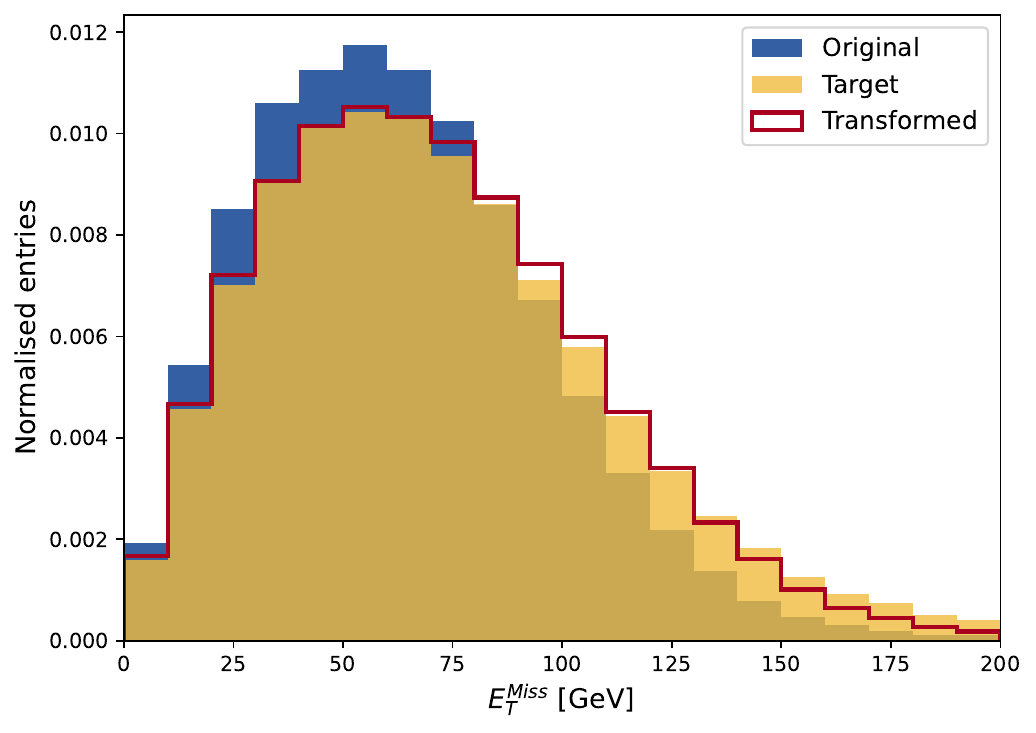}  \includegraphics[width=0.32\textwidth]{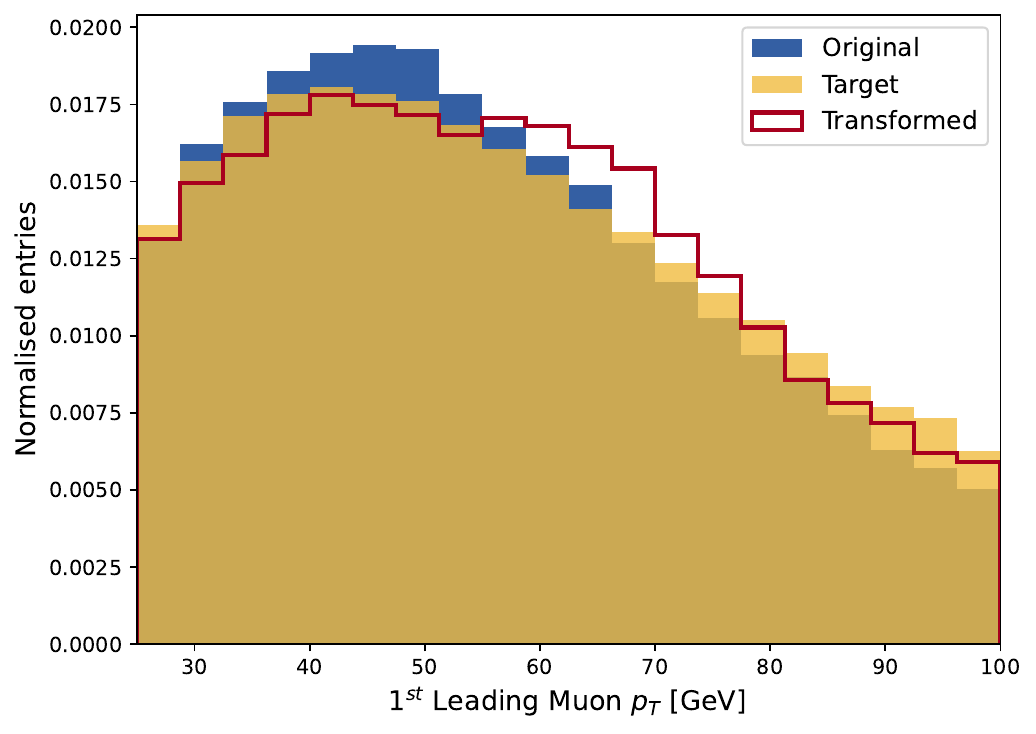}
    \includegraphics[width=0.32\textwidth]{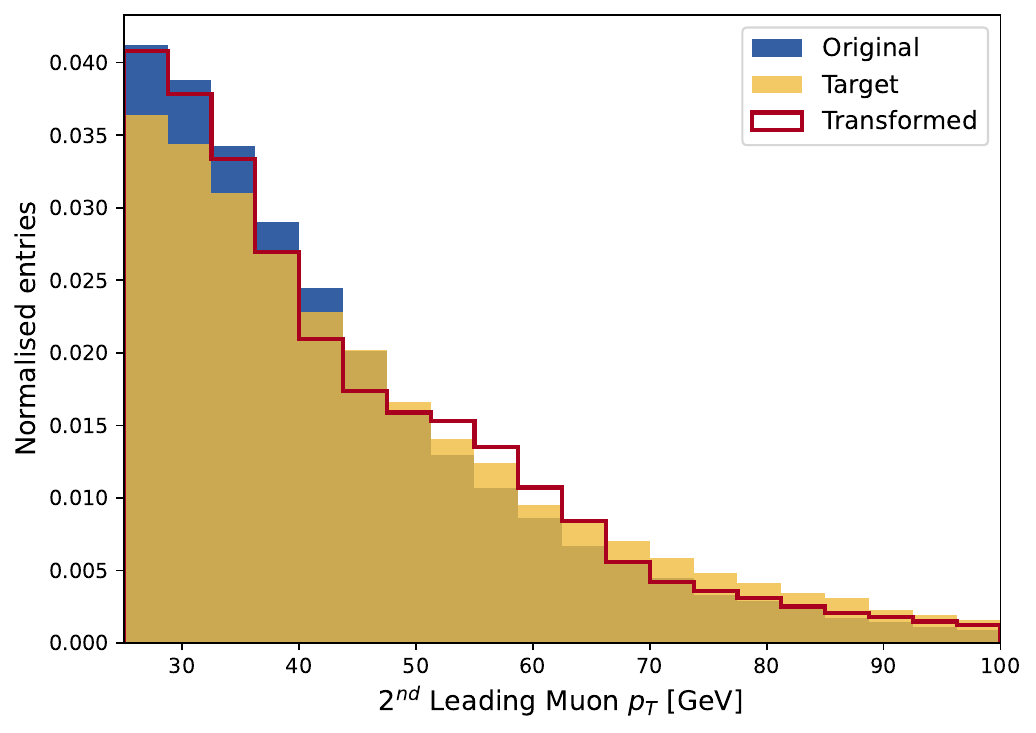}
    \includegraphics[width=0.32\textwidth]{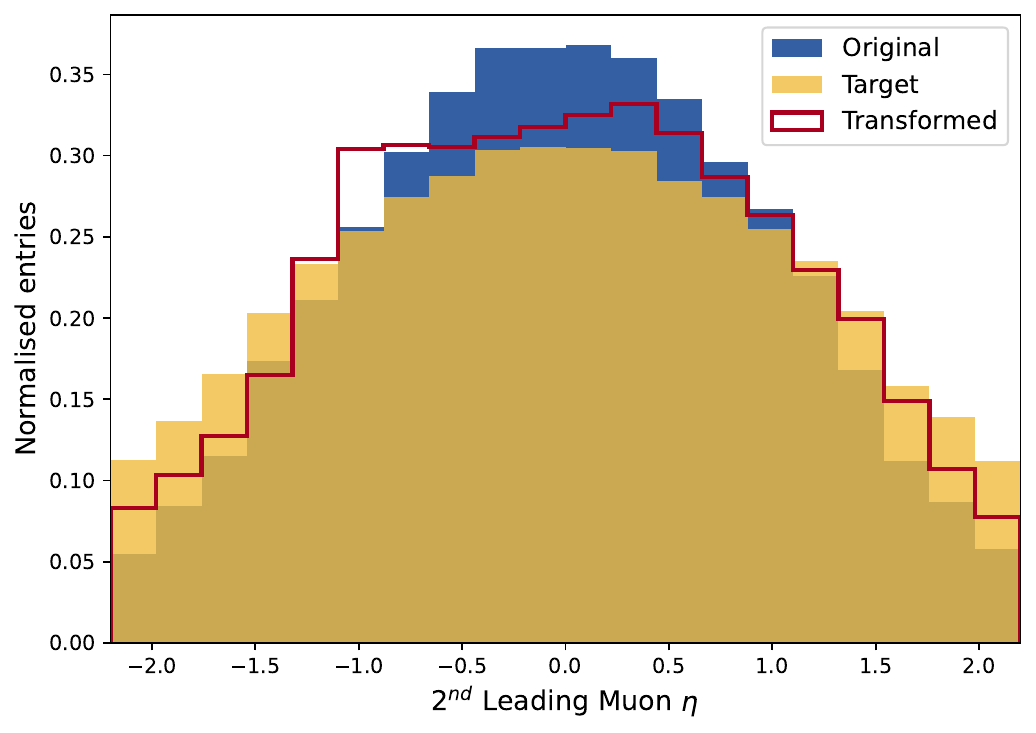}
    \includegraphics[width=0.32\textwidth]{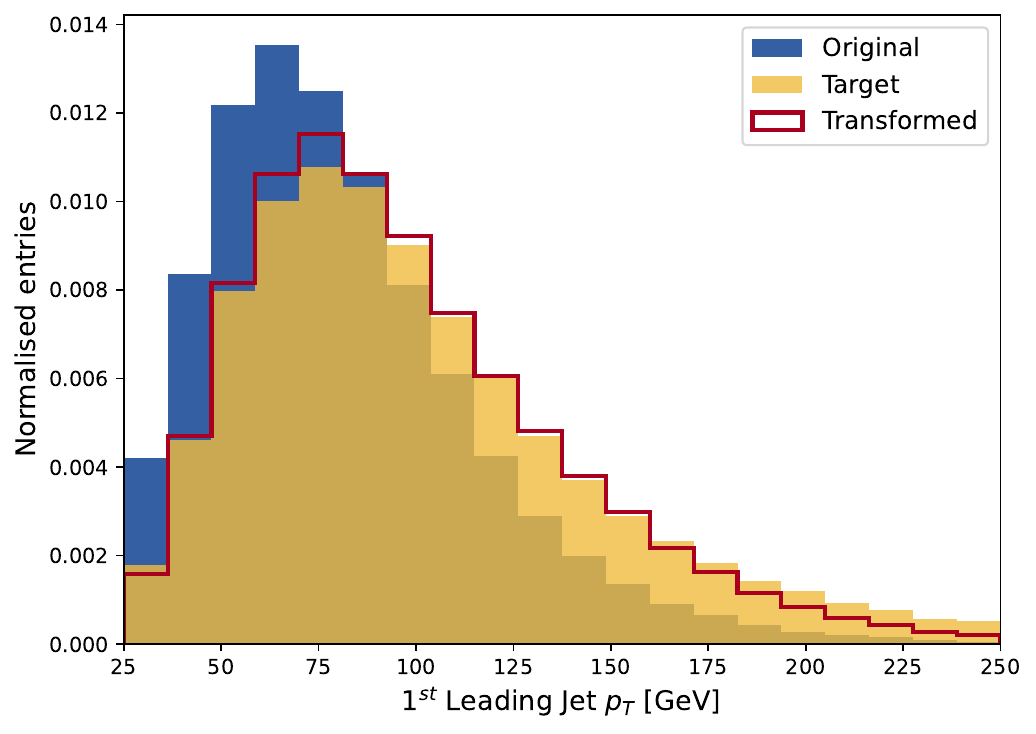}
    \includegraphics[width=0.32\textwidth]{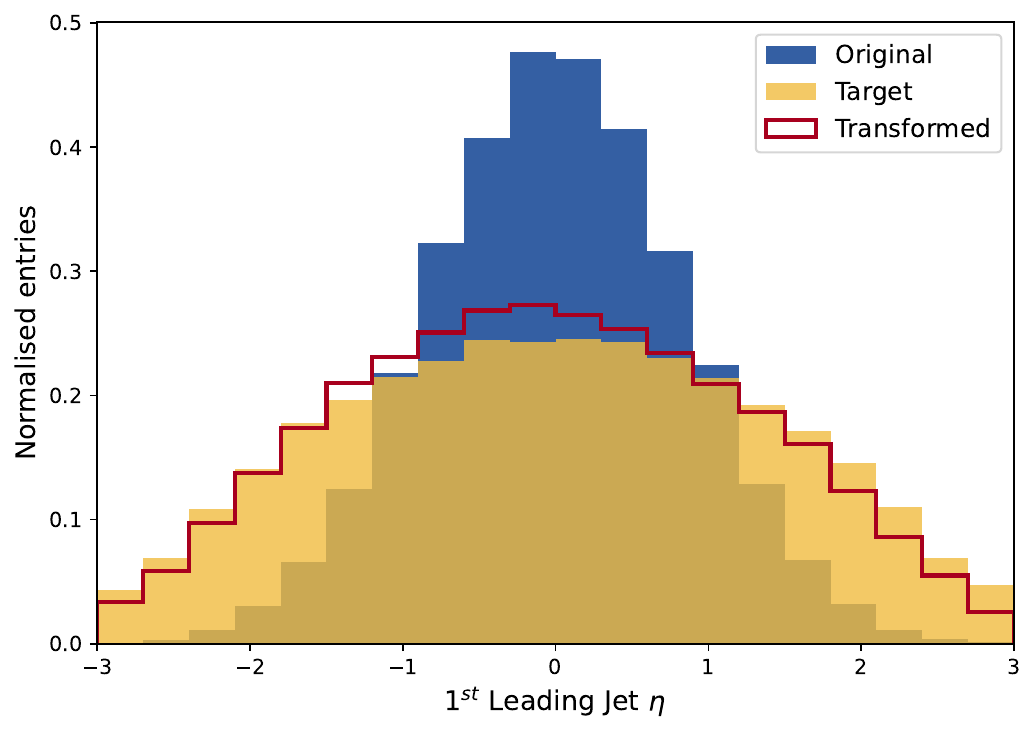}
    \caption{Comparison of selected input features for the underlying event models of the original data-set (blue), the target data-set (yellow) and the transformed results, predicted by the neural network (violet)}
    \label{fig:UEFeaturesTTBarGlobal}
\end{figure}

Figure~\ref{fig:UEFeaturesTTBarGlobal} shows the comparison of selected input feature distributions for the original sample, the target sample, and the transformed output. Overall, a good level of agreement is achieved, indicating that the network successfully learns the dominant shifts in the kinematic distributions. However, for certain observables, such as the pseudorapidity of the second muon, noticeable discrepancies remain after training. The remaining discrepancies may partly originate from the bounded residual parameterization introduced in Eq.~(\ref{eqn:34alt}). Since the hyperbolic tangent restricts the correction to $|\Delta_{\theta,j}(x)| \leq \alpha_j$, the parameter $\alpha_j$ directly sets the maximum displacement allowed for feature $j$. This constraint is introduced deliberately to prevent large, potentially unphysical modifications of the baseline simulation. However, if the difference between the original and target distributions requires corrections larger than the chosen $\alpha_j$, the transformation cannot achieve exact closure, even if the neural network itself has sufficient expressive capacity.

\begin{figure}
    \centering
    \includegraphics[width=0.32\textwidth]{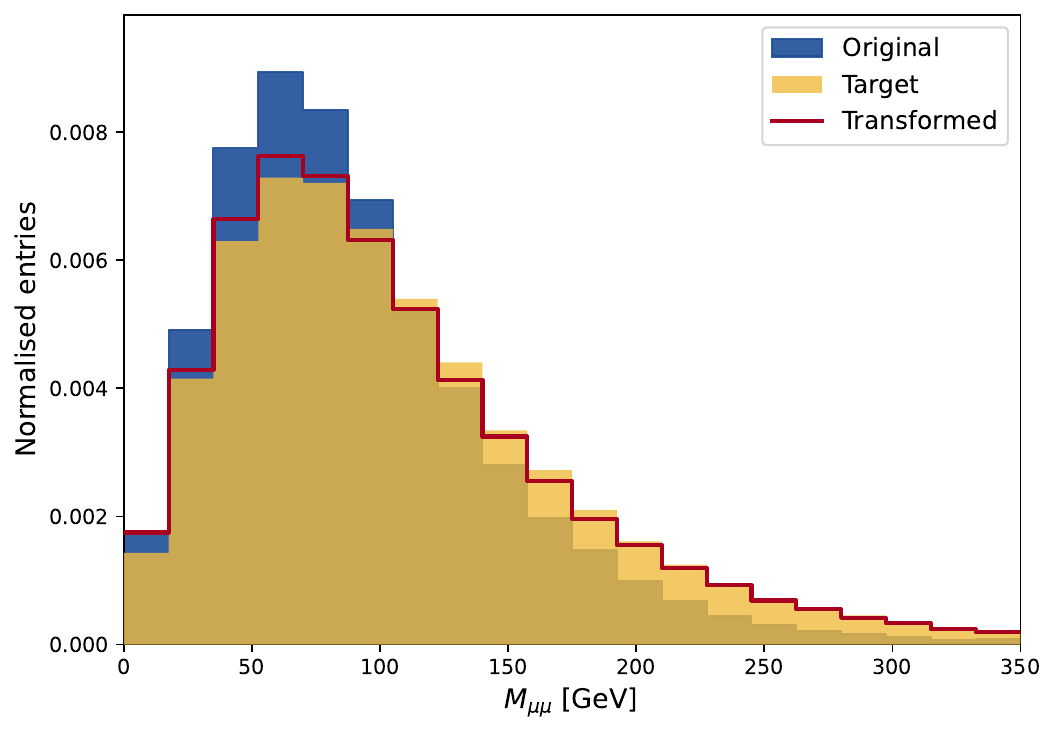}
    \includegraphics[width=0.32\textwidth]{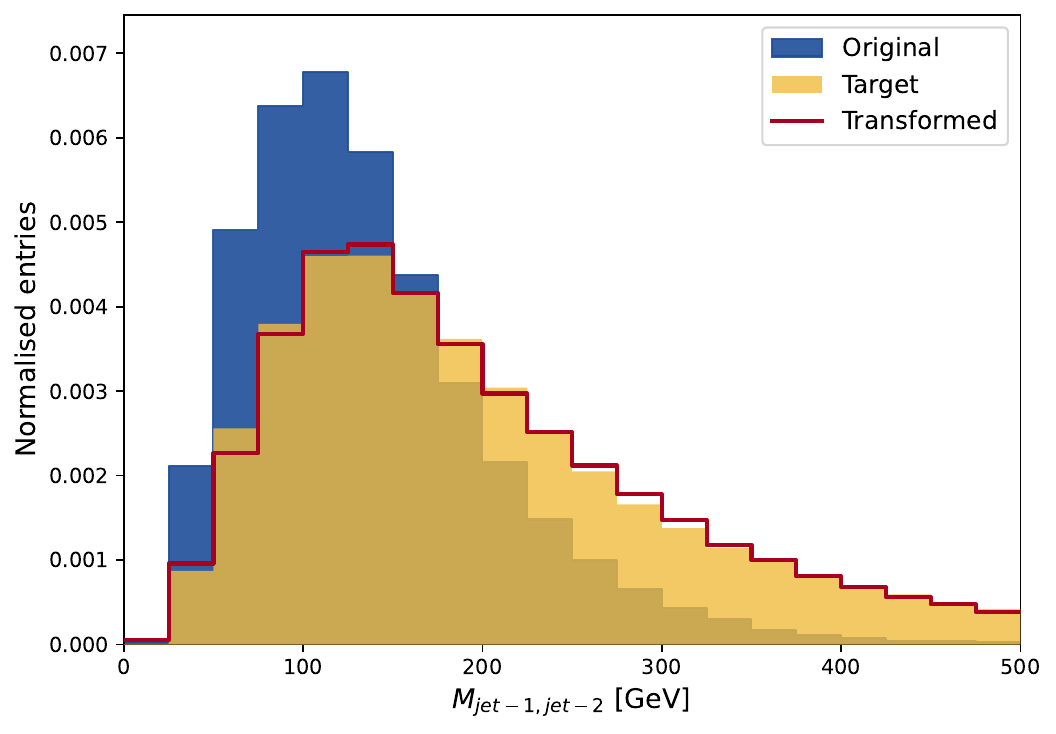}
    \includegraphics[width=0.32\textwidth]{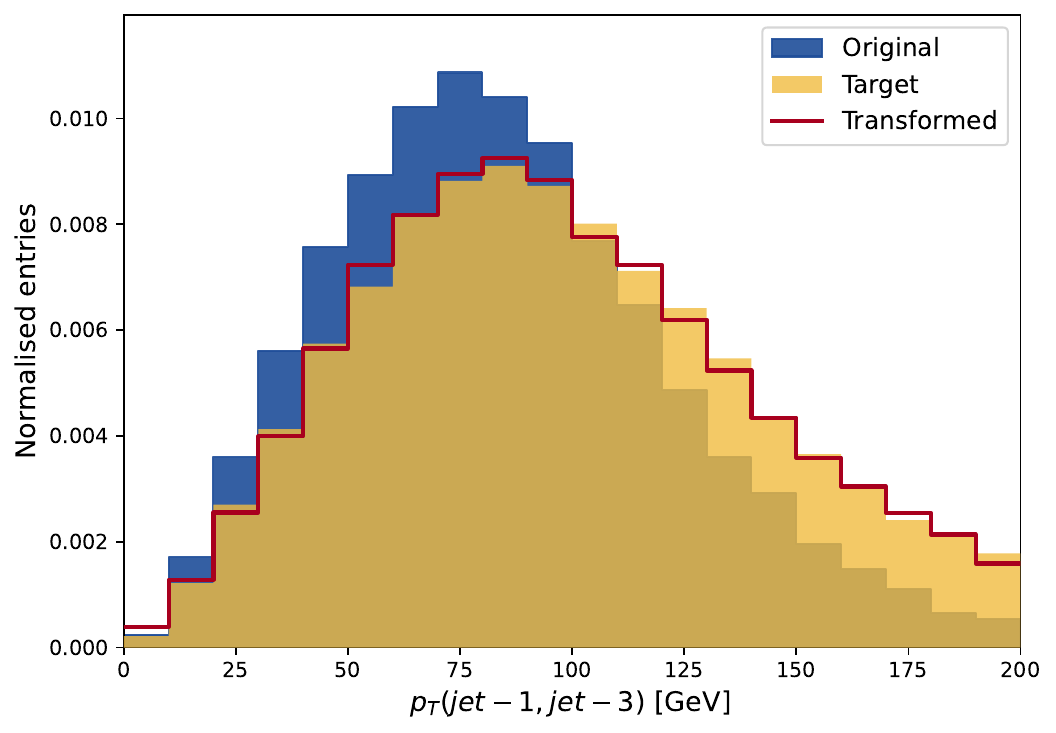}
    \caption{Comparison of derived observables based in the input features for the underlying event models of the original data-set (blue), the target data-set (yellow) and the transformed results, predicted by the neural network (violet)}
    \label{fig:UEDerivedQuantitiesTTBarGlobal}
\end{figure}

Figure~\ref{fig:UEDerivedQuantitiesTTBarGlobal} presents the corresponding comparison for derived observables, including the invariant mass of the muon pair, the invariant mass of the two leading jets, and the transverse momentum of a multi-jet system. For these quantities, an excellent level of closure is observed, demonstrating that the model is capable of capturing non-trivial correlations between input features, even though the training is primarily driven by one-dimensional distributions.

\begin{figure}
    \centering
    \includegraphics[width=0.32\textwidth]{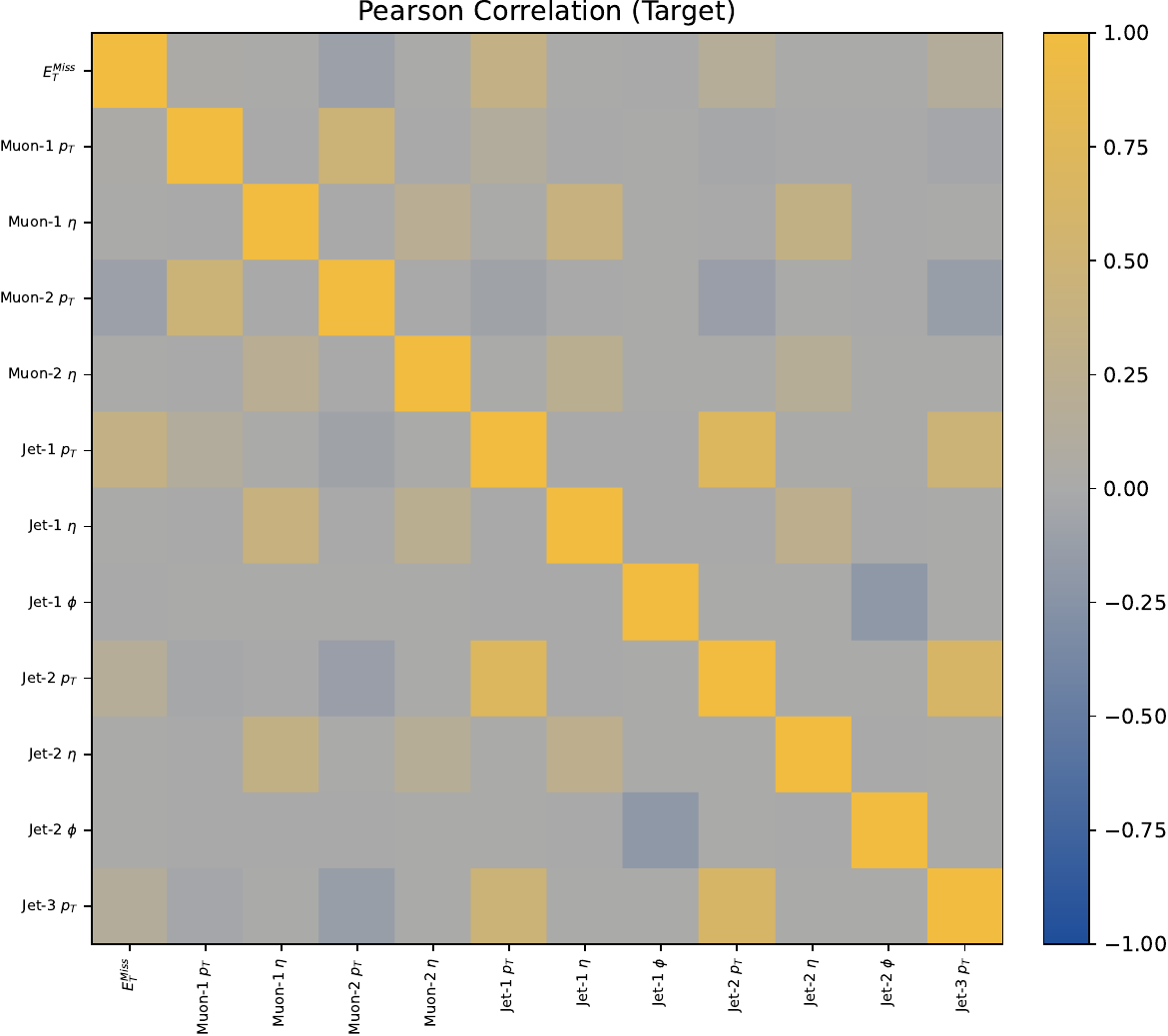}
    \includegraphics[width=0.32\textwidth]{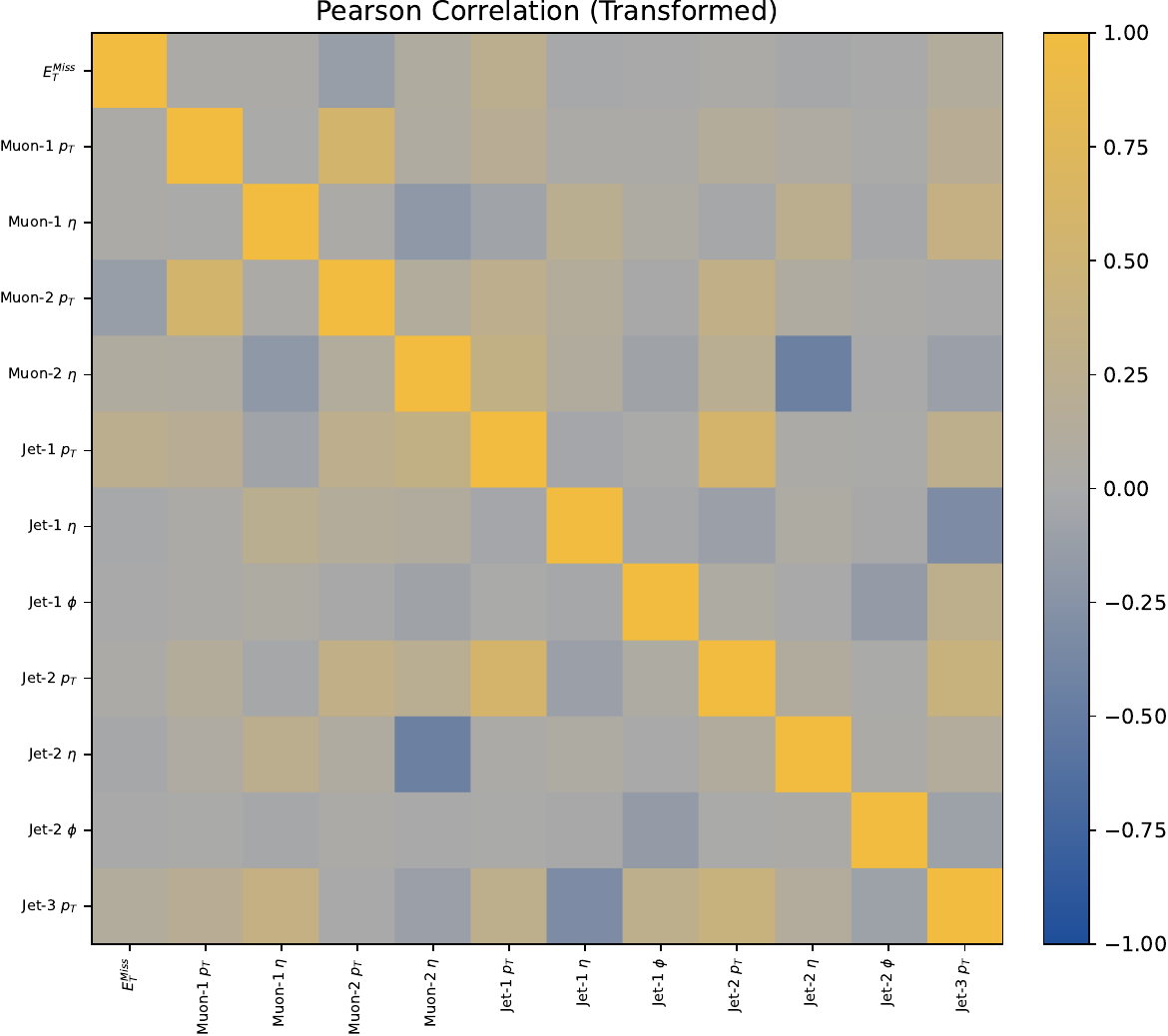}
    \includegraphics[width=0.32\textwidth]{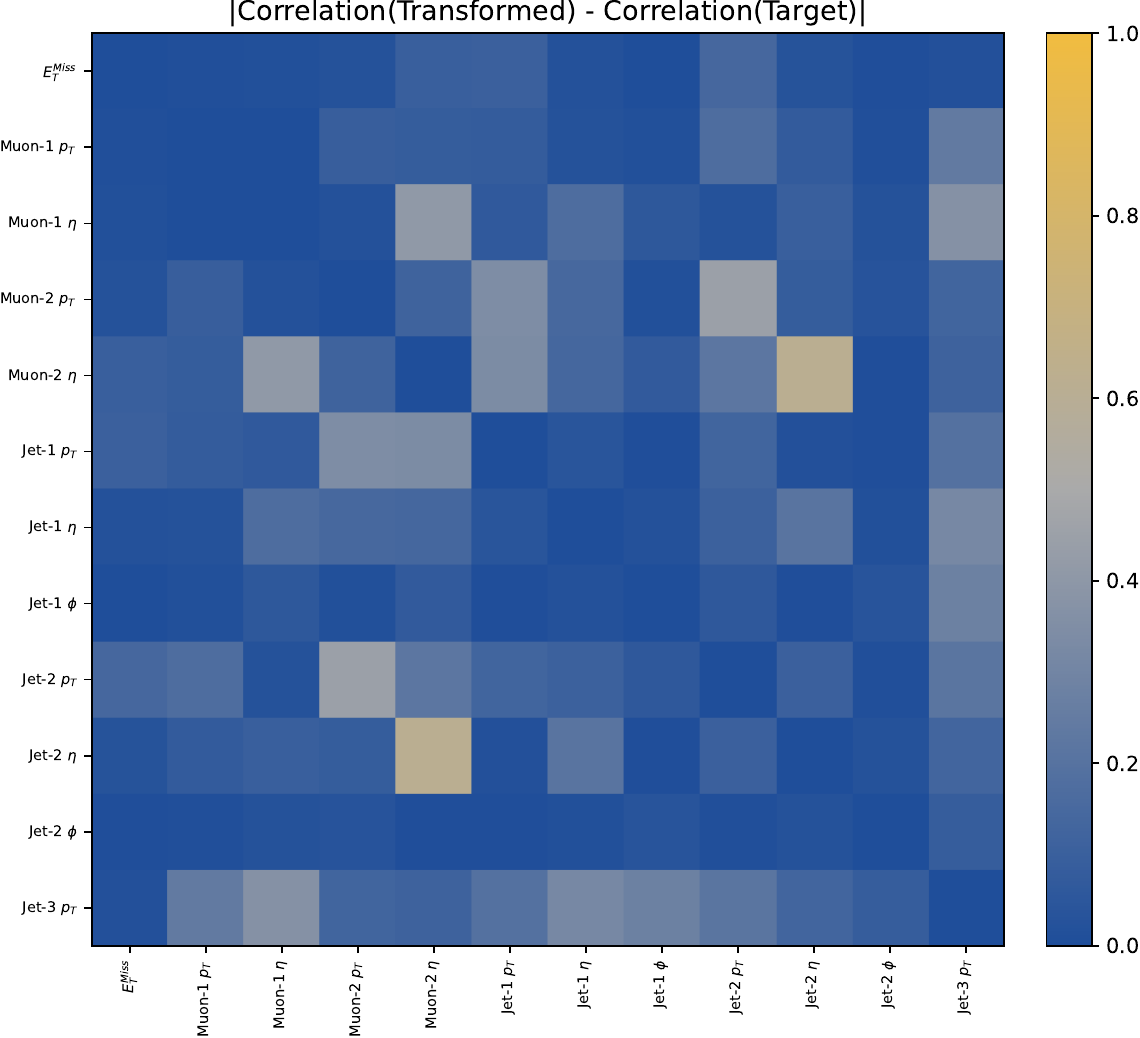}
    \caption{Comparison of Pearson correlation coefficients of the target data-set (left), the transformed results by the neural network (middle) and their difference (right) for the Global TTBar event models.}
    \label{fig:UECorrelationTTBarGlobal}
\end{figure}

Finally, the agreement in the correlation structure between the transformed
and target samples is quantified using the Pearson correlation matrix, as
shown in Figure~\ref{fig:UECorrelationTTBarGlobal}. The left and middle
panels show the correlation matrices of the target and transformed samples,
respectively, while the right panel shows their difference. The average
deviation is found to be at the level of $\mathcal{O}(0.2)$, with the largest
discrepancies reaching up to $\sim 0.5$ for correlations involving jet
kinematics. This indicates that, while the global model captures a significant
fraction of the multivariate structure, certain complex dependencies remain
challenging to reproduce fully.

\subsection{Transfer from Top Quark Pair Production Processes from Tevatron to the LHC via a Two-Step Residual Transformation}

We now revisit the same $t\bar{t}$ transfer task studied in the previous section, but apply the two-step residual transformation instead of the global approach, keeping the overall training setup unchanged. In this formulation, the transformation is decomposed into two stages. In the first stage, independent feature-wise residual networks are trained to match the one-dimensional distributions of individual input features. Each network predicts a bounded correction for a single feature, using a subset of inputs as context, thereby focusing on reproducing marginal distributions while keeping the modifications localized and controlled.

\begin{figure}
    \centering
    \includegraphics[width=0.32\textwidth]{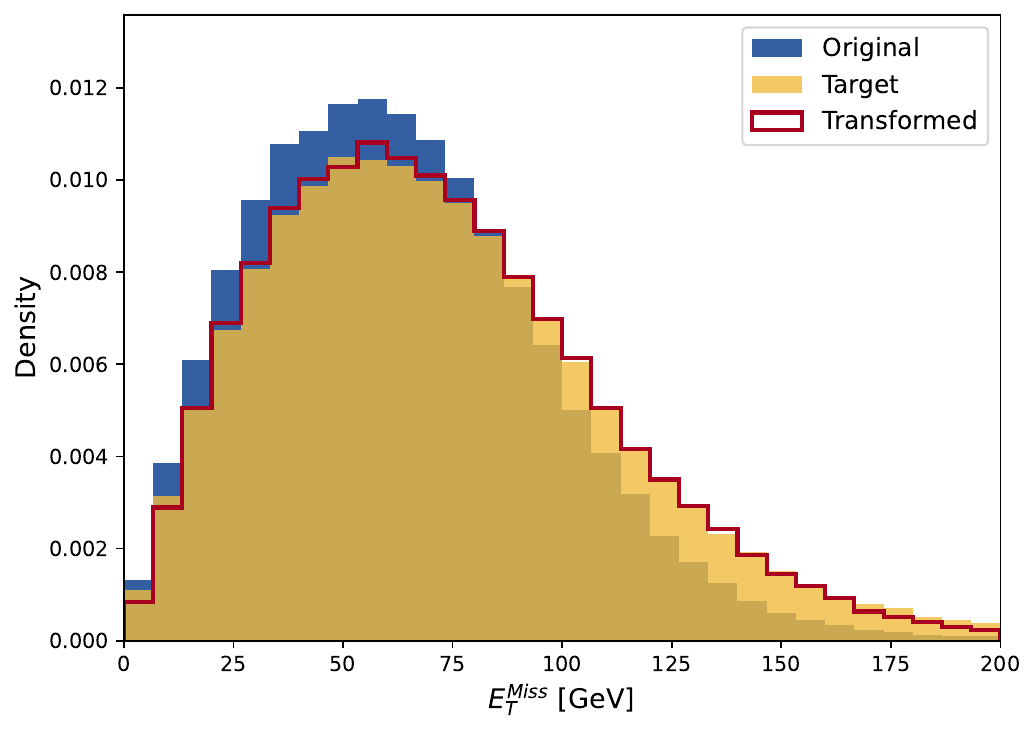}
    \includegraphics[width=0.32\textwidth]{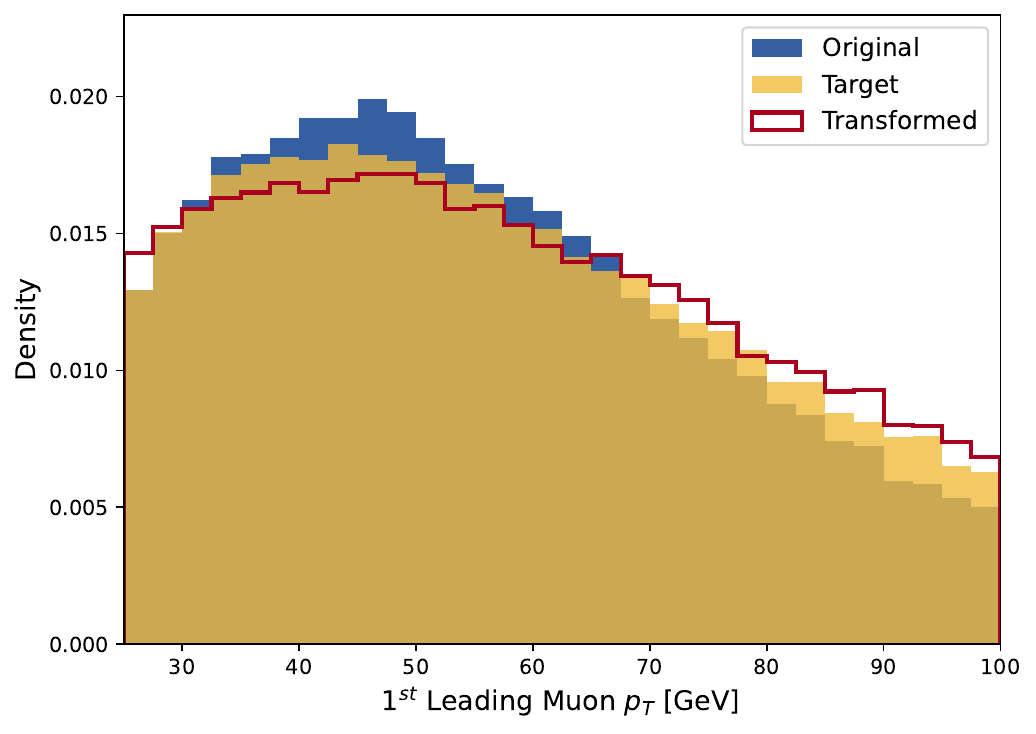}
    \includegraphics[width=0.32\textwidth]{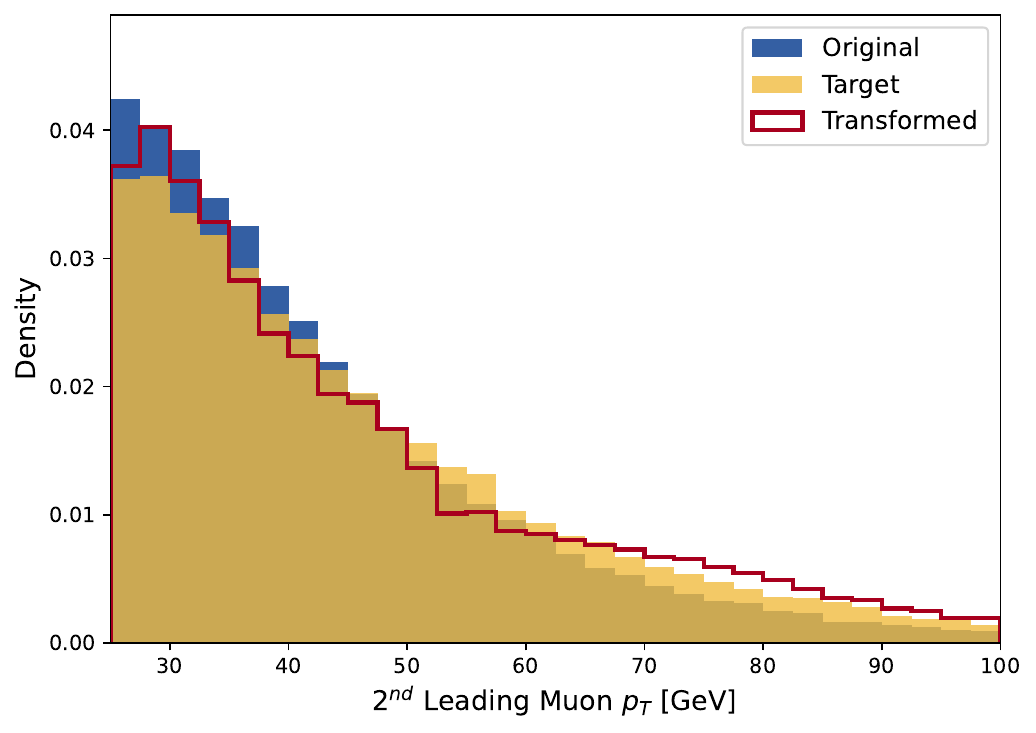}
    \includegraphics[width=0.32\textwidth]{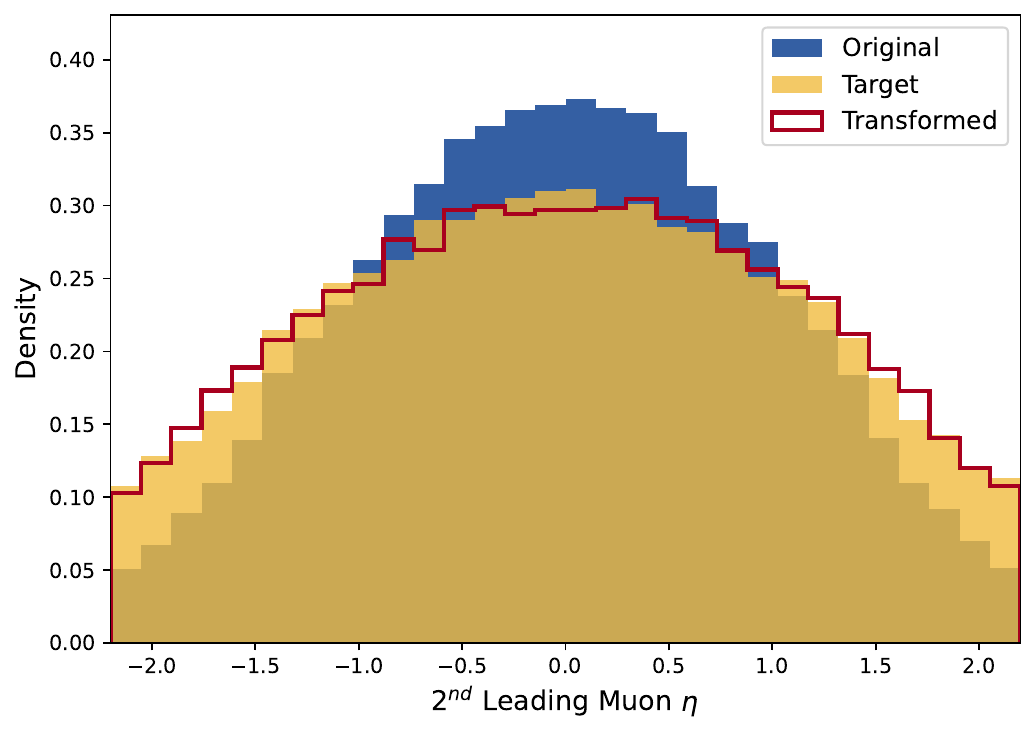}
    \includegraphics[width=0.32\textwidth]{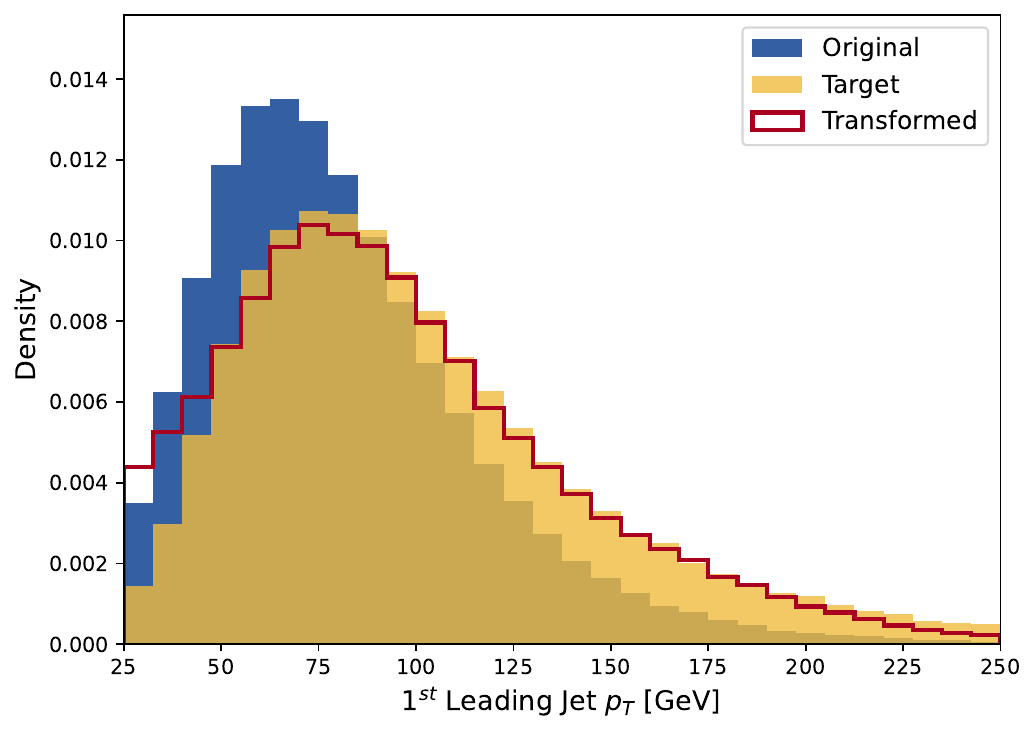}
    \includegraphics[width=0.32\textwidth]{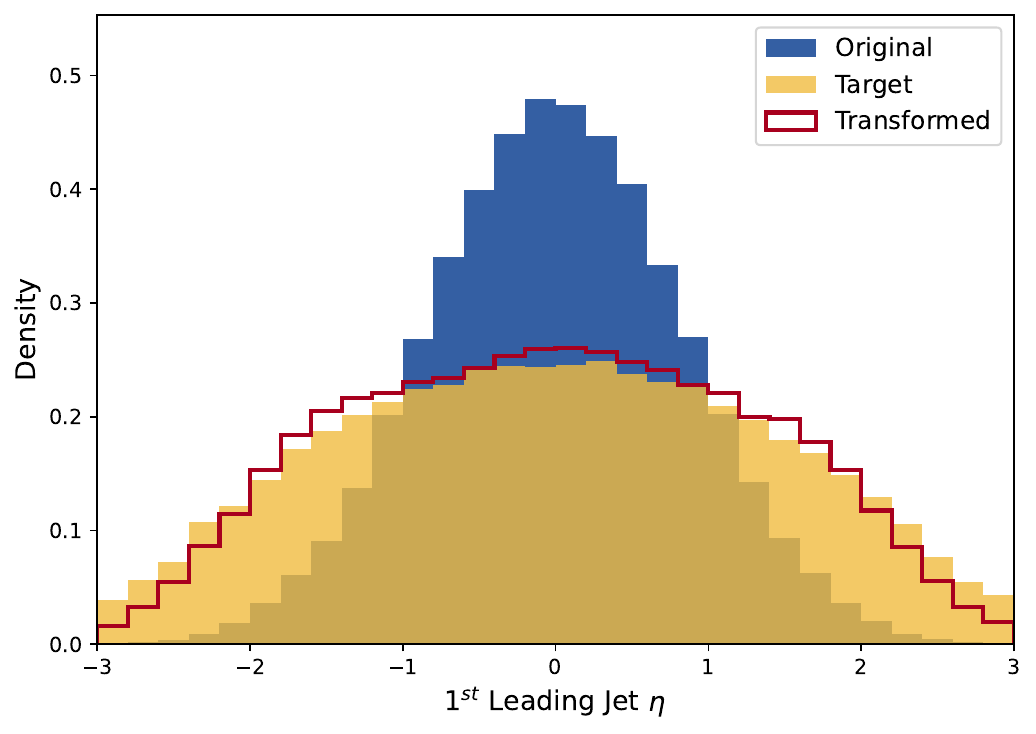}
    \caption{Comparison of selected input features for the underlying event models of the original data-set (blue), the target data-set (yellow) and the transformed results, predicted by the neural network (violet)}
    \label{fig:UEFeaturesTTBarIndividual}
\end{figure}

In the second stage, a global residual network is applied to the output of the first step. This refinement network operates on the full feature vector and is trained to improve agreement in derived observables and to restore correlations between features. In this way, the two-step approach explicitly separates the learning of one-dimensional distributions from the modeling of multivariate structure.

\begin{figure}
    \centering
    \includegraphics[width=0.32\textwidth]{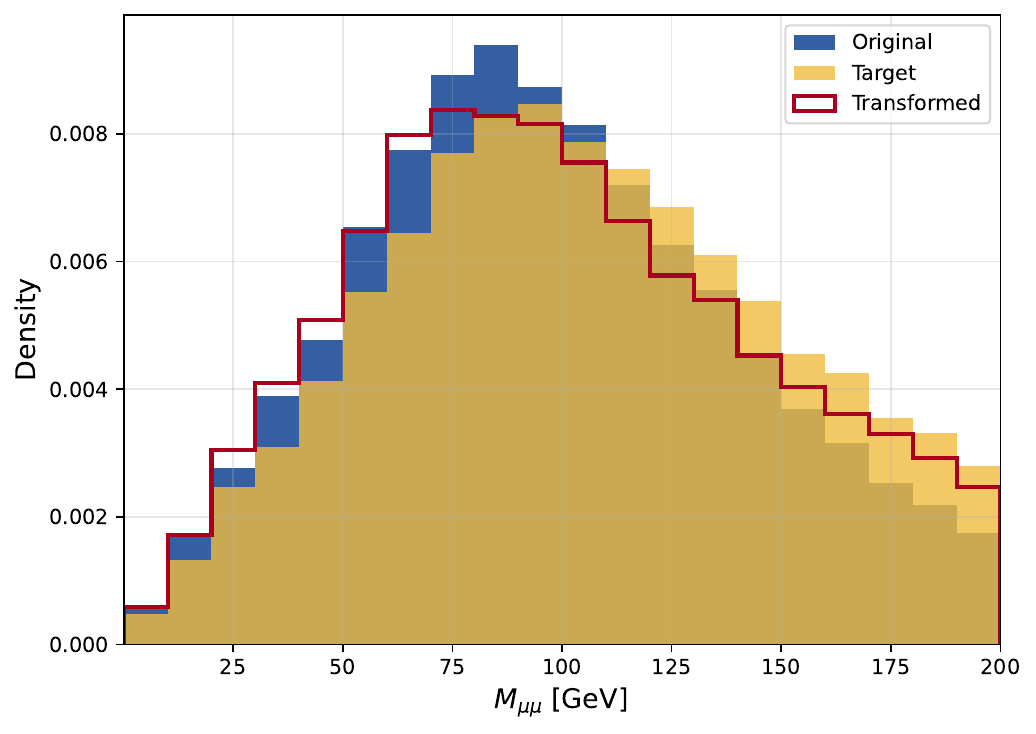}
    \includegraphics[width=0.32\textwidth]{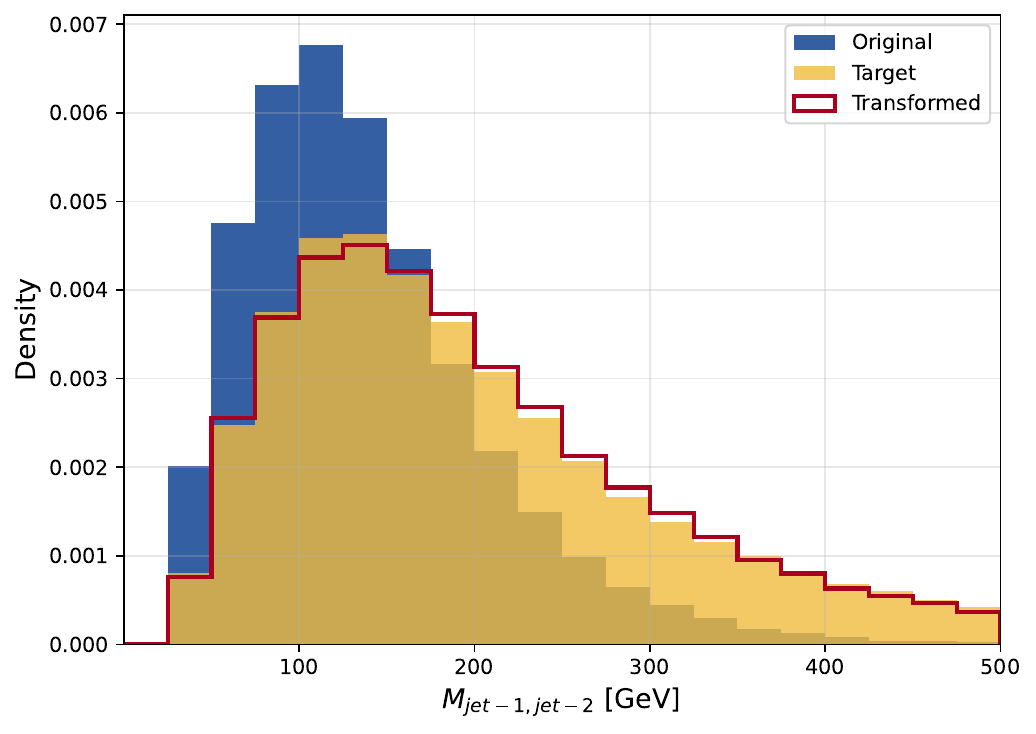}
    \includegraphics[width=0.32\textwidth]{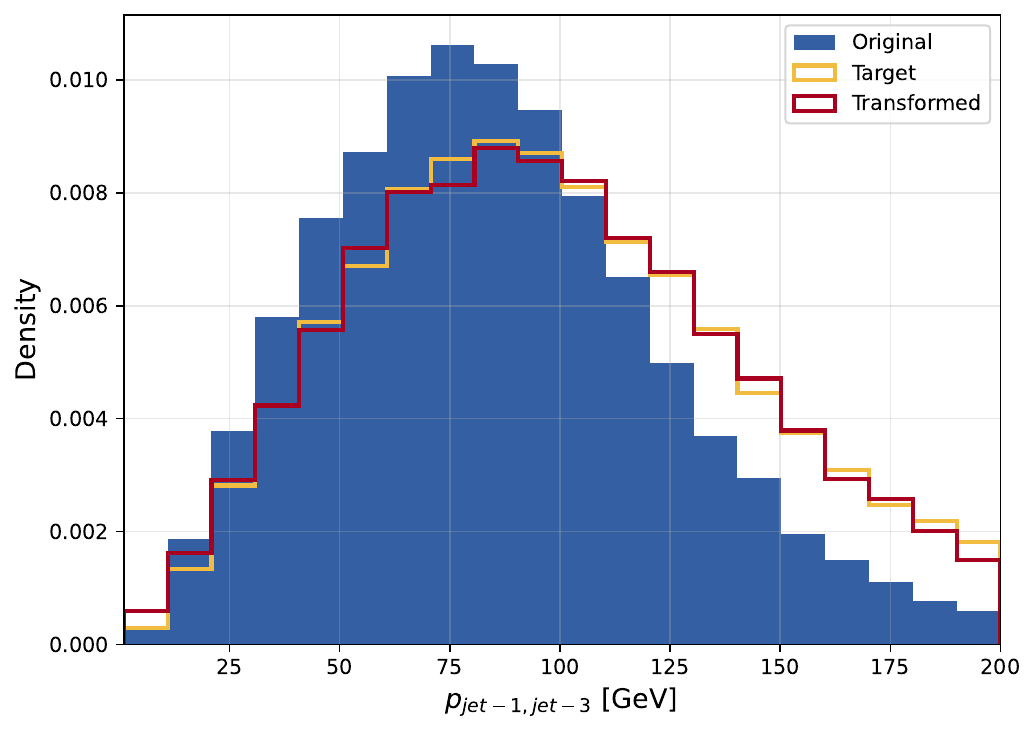}
    \caption{Comparison of derived observables based in the input features for the underlying event models of the original data-set (blue), the target data-set (yellow) and the transformed results, predicted by the neural network (violet)}
    \label{fig:UEDerivedQuantitiesTTBarIndividual}
\end{figure}

The resulting feature distributions are shown in Figure~\ref{fig:UEFeaturesTTBarIndividual}, where the original, target, and transformed samples are compared for a set of representative observables. Compared to the global approach, an improved level of agreement is observed across most features, indicating that the dedicated feature-wise training provides a more precise control of individual distributions. Figure~\ref{fig:UEDerivedQuantitiesTTBarIndividual} shows the corresponding comparison for derived observables. A very good level of agreement is achieved for jet-related quantities, demonstrating that the second-stage refinement successfully restores relevant correlations. However, small residual discrepancies remain in the invariant mass distribution of the muon pair, suggesting that certain correlated structures are still not fully captured.

\begin{figure}
    \centering
    \includegraphics[width=0.32\textwidth]{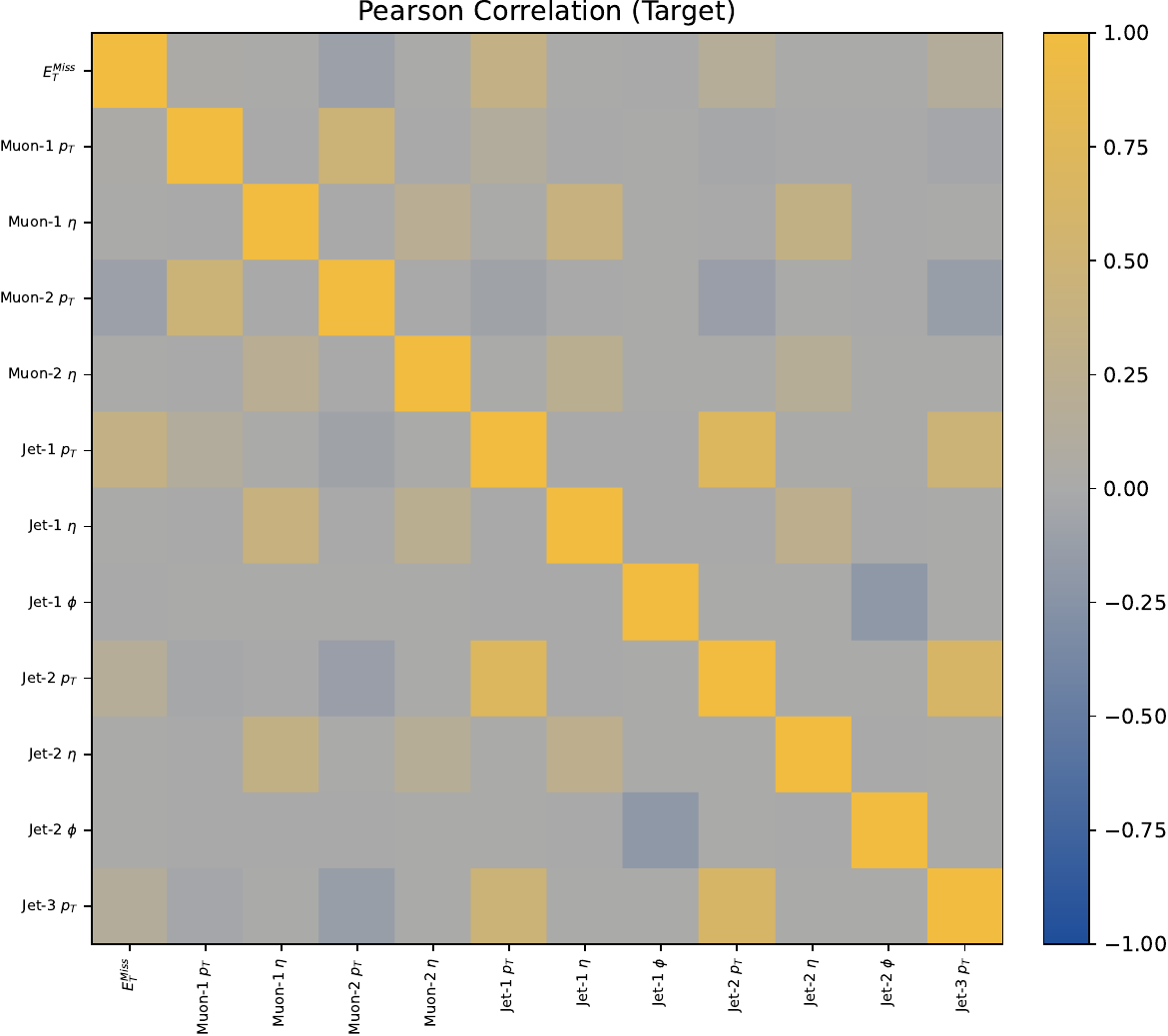}
    \includegraphics[width=0.32\textwidth]{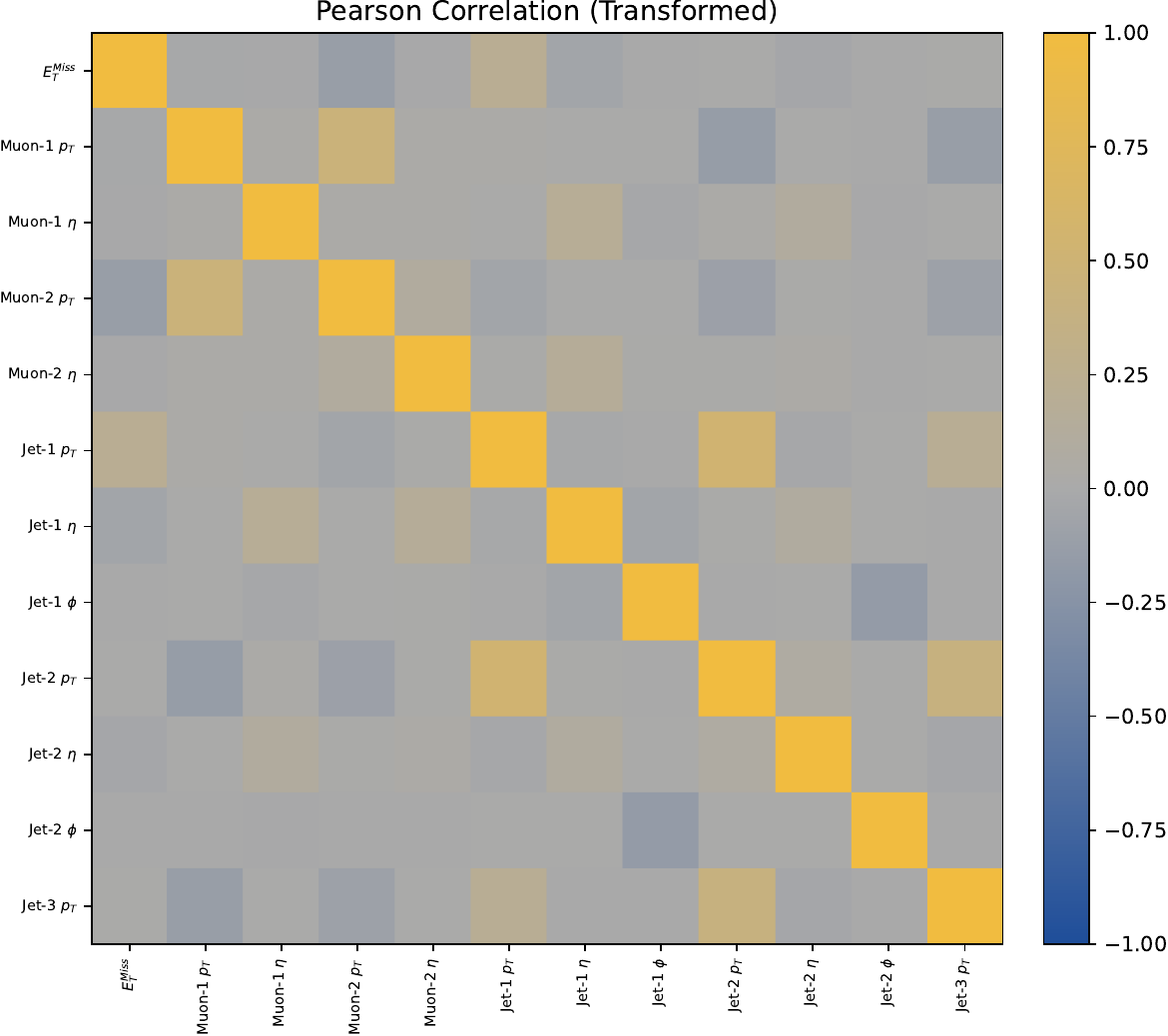}
    \includegraphics[width=0.32\textwidth]{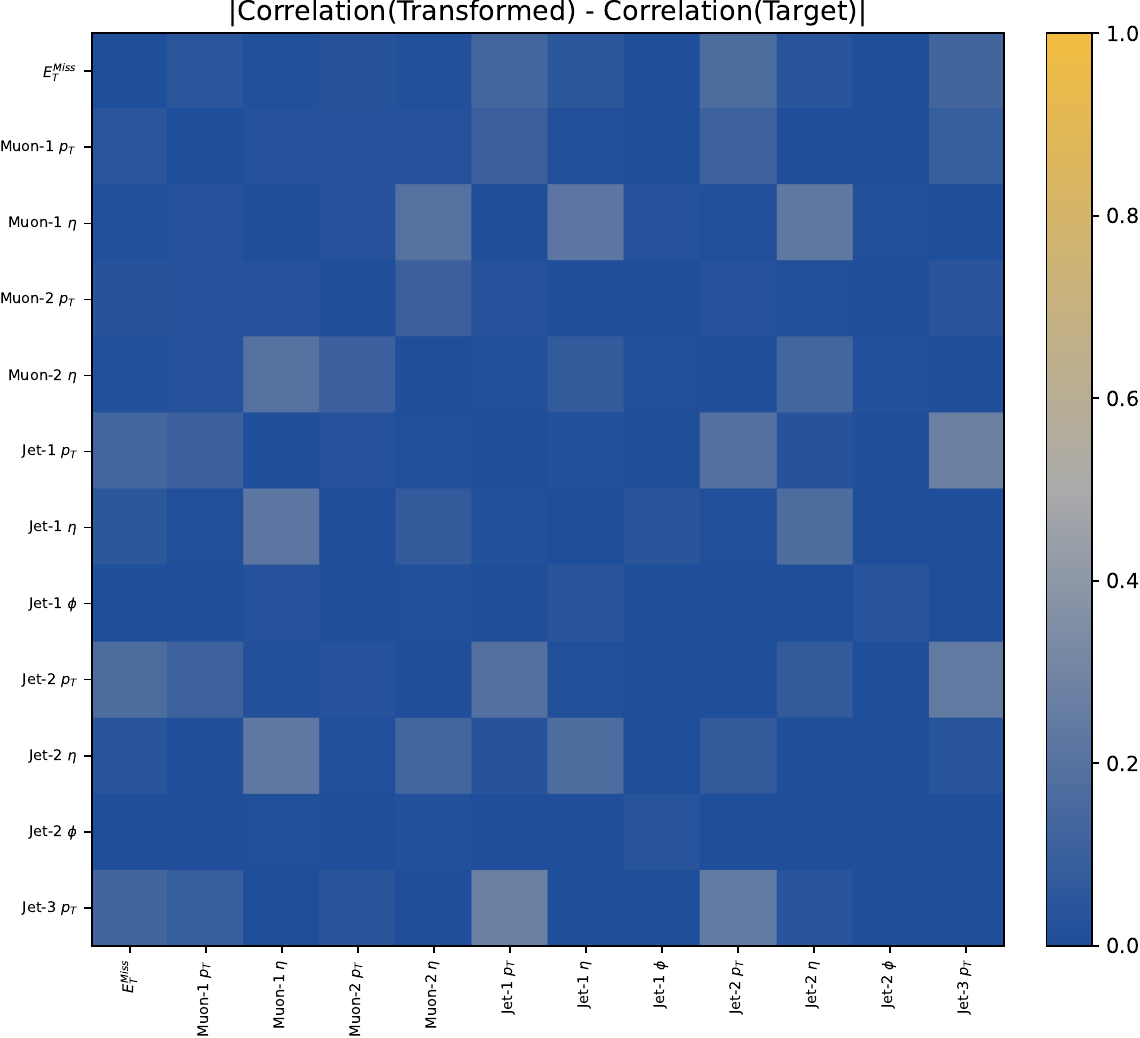}
    \caption{Comparison of Pearson correlation coefficients of the target data-set (left), the transformed results by the neural network (middle) and their difference (right) for the underlying event models.}
    \label{fig:UECorrelationTTBar}
\end{figure}

The preservation of correlations is illustrated in Figure~\ref{fig:UECorrelationTTBar}. The differences between the transformed and target Pearson correlation matrices are consistently below 0.2, representing a clear improvement over the global approach. This demonstrates that the two-step procedure is effective in maintaining the multivariate structure while achieving high accuracy at the level of individual feature distributions.

\subsection{Transfer between different Underlying Event Models via a Two-Step Residual Transformation}

We now consider a more complex transfer between different underlying event (UE) models, focusing on the transformation from the AZ tune to the Monash tune. In this setup, each event is described by 90 input features, corresponding to the kinematic properties of the 30 leading charged tracks, namely their transverse momentum ($p_T$), pseudorapidity ($\eta$), and azimuthal angle ($\phi$).
The training setup follows the same procedure as before, with 80\% of the data used for training and 20\% for validation, and early stopping applied based on the validation loss. The batch-size was chosen again to be larger than 5k events, ensuring the stabilty of the histogram related losses.

\begin{figure}
    \centering
    \includegraphics[width=0.32\textwidth]{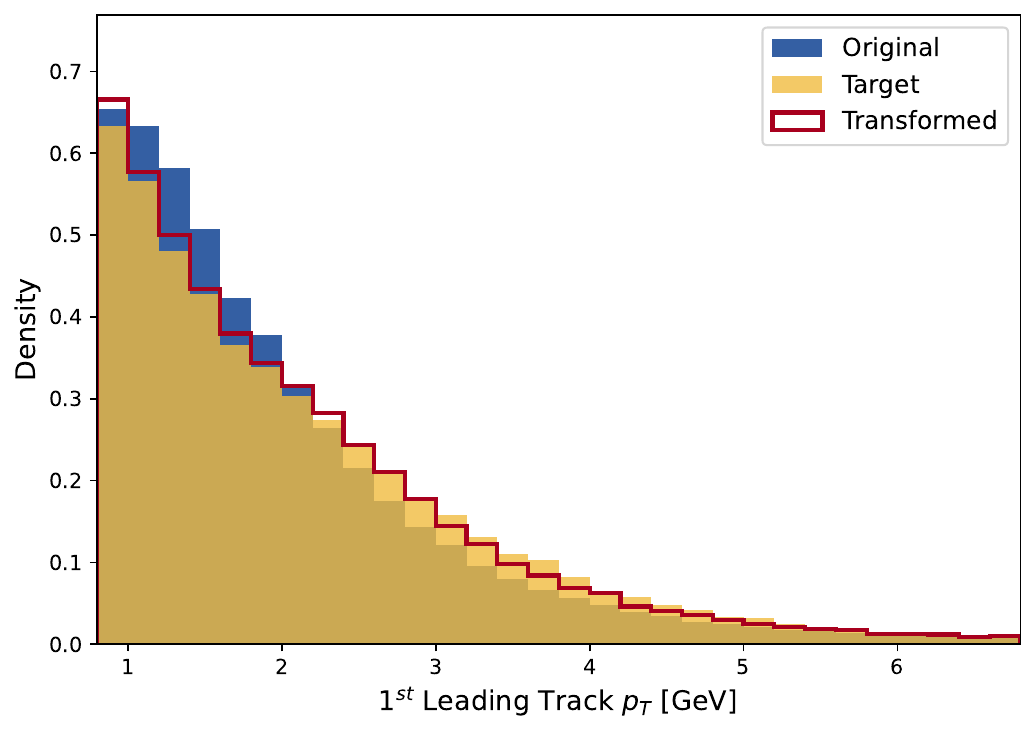}
    \includegraphics[width=0.32\textwidth]{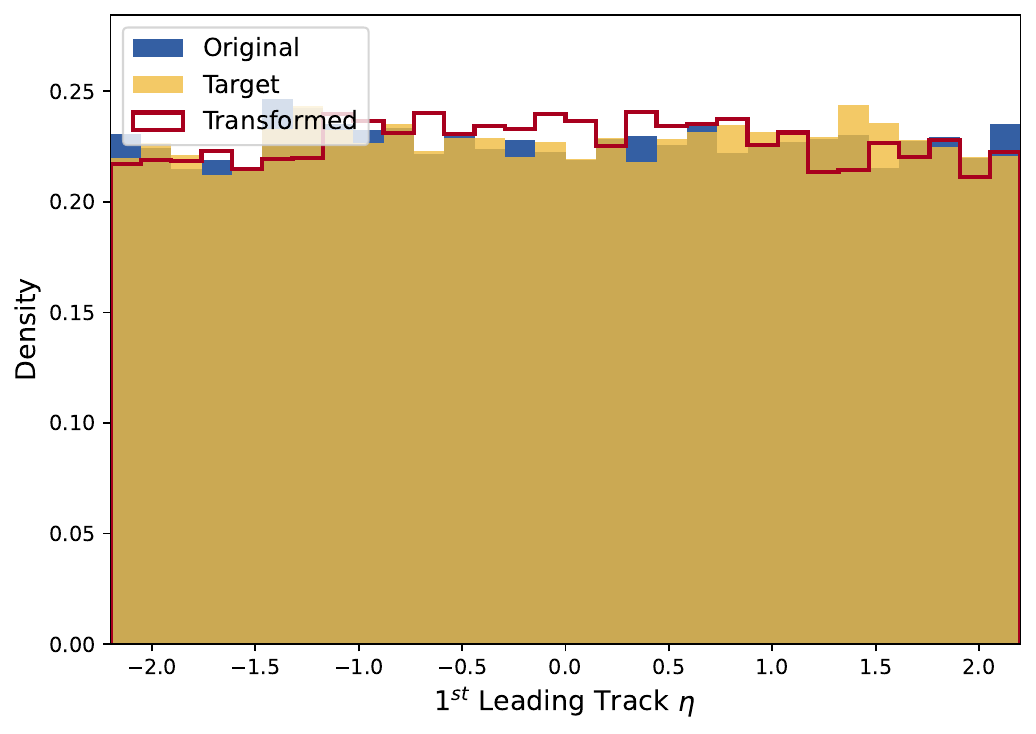}
    \includegraphics[width=0.32\textwidth]{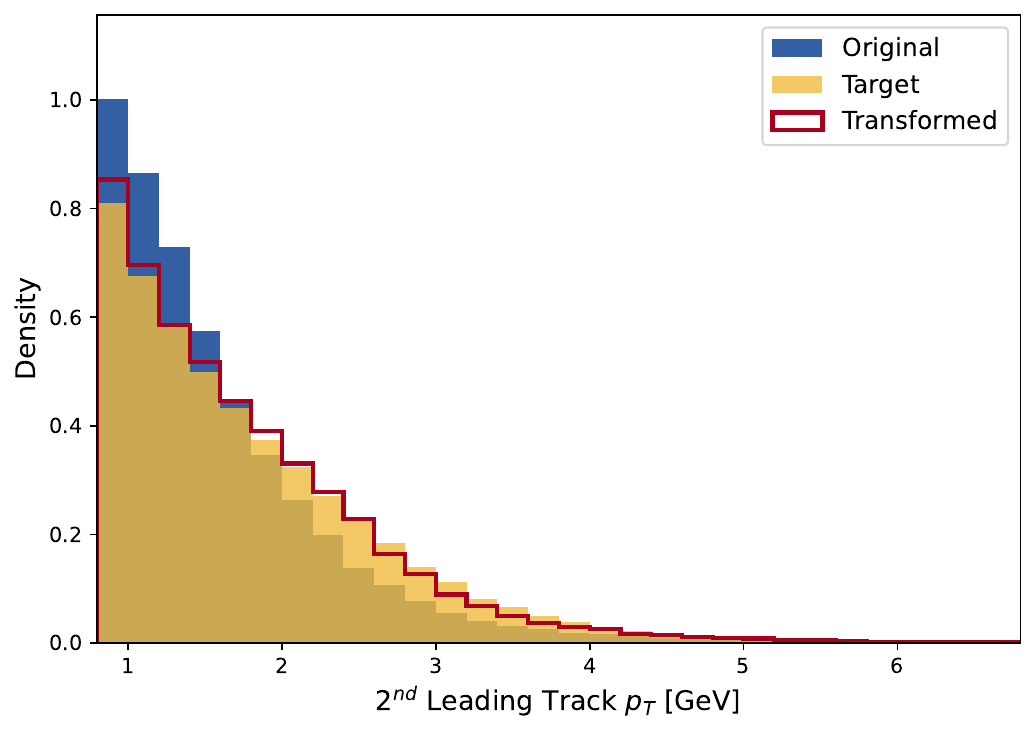}
    \includegraphics[width=0.32\textwidth]{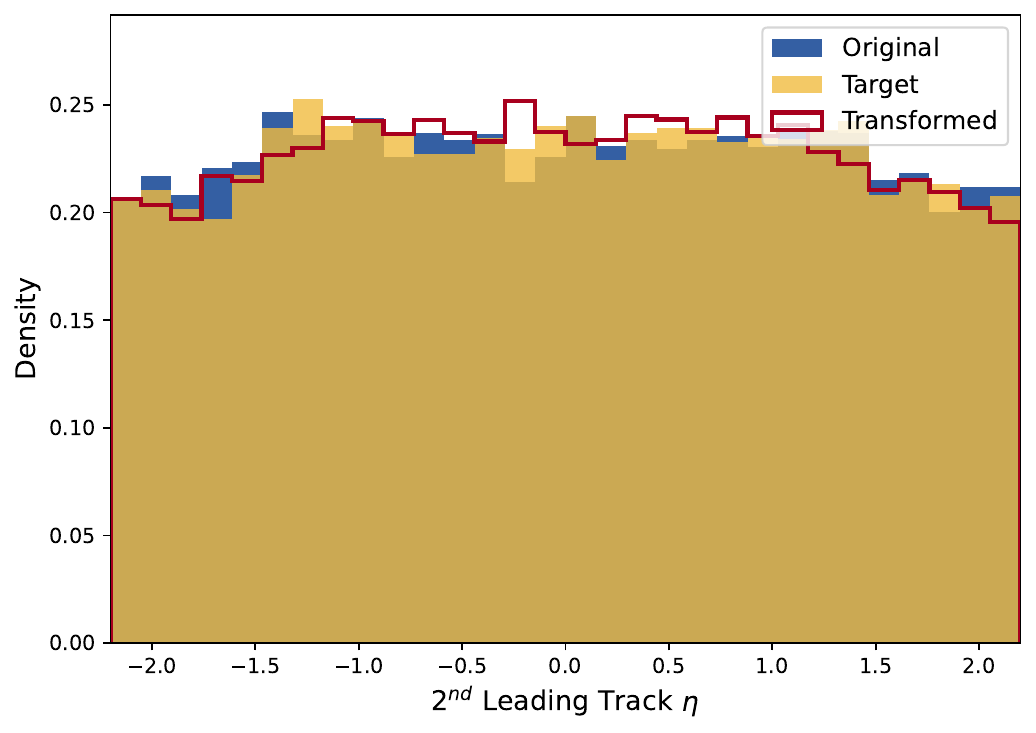}
    \includegraphics[width=0.32\textwidth]{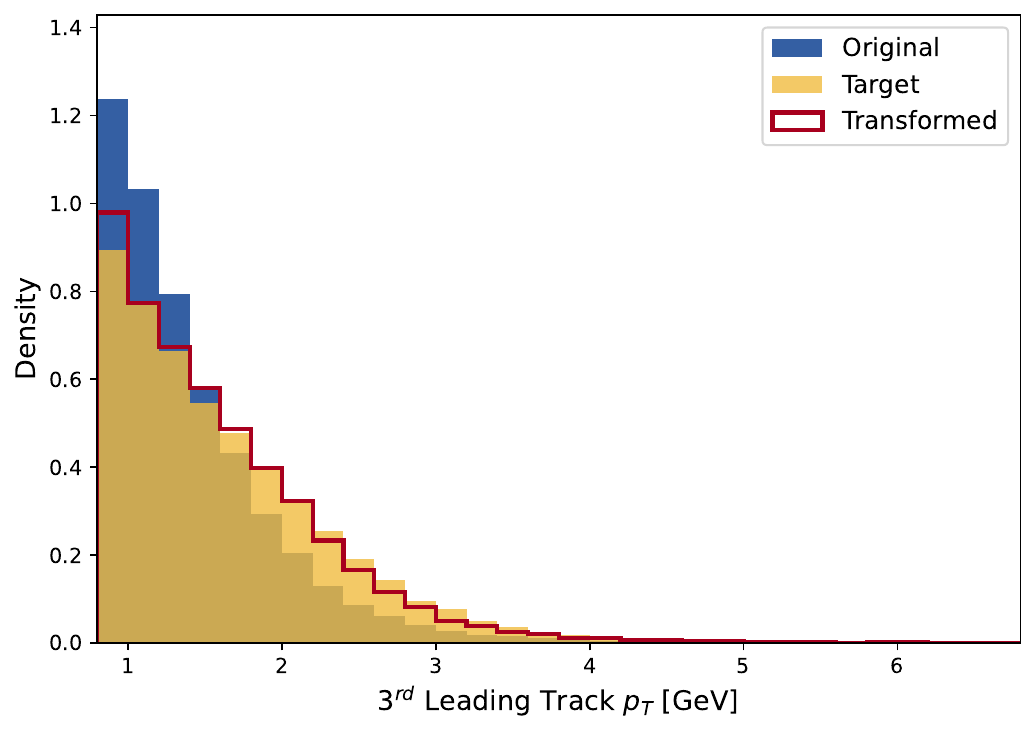}
    \includegraphics[width=0.32\textwidth]{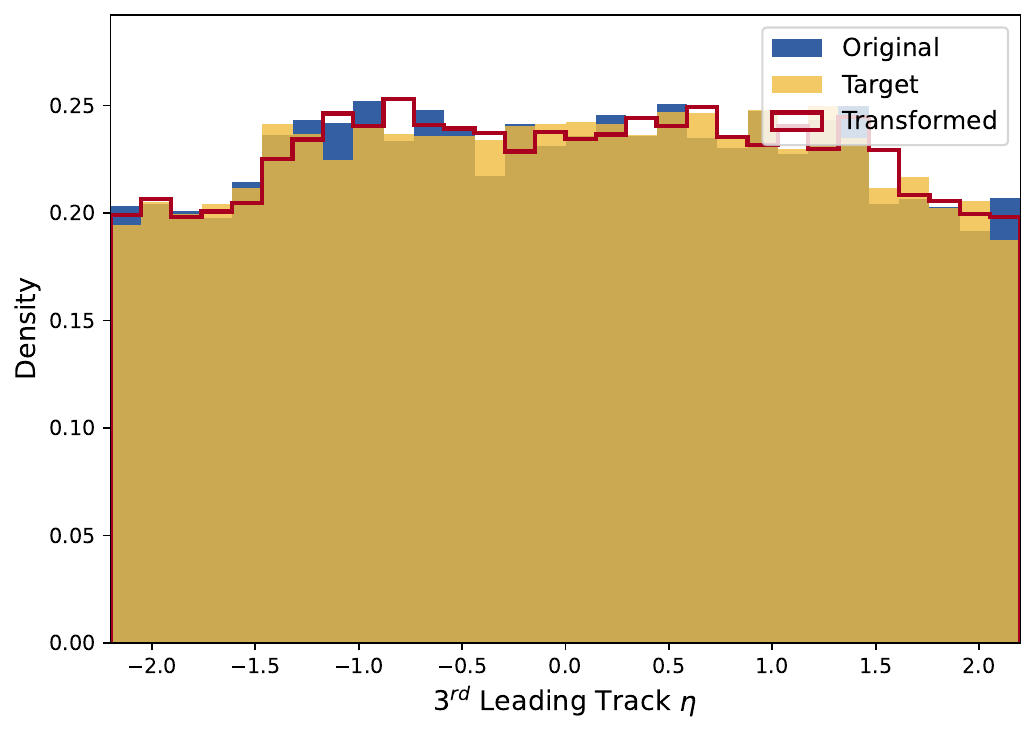}
    \includegraphics[width=0.32\textwidth]{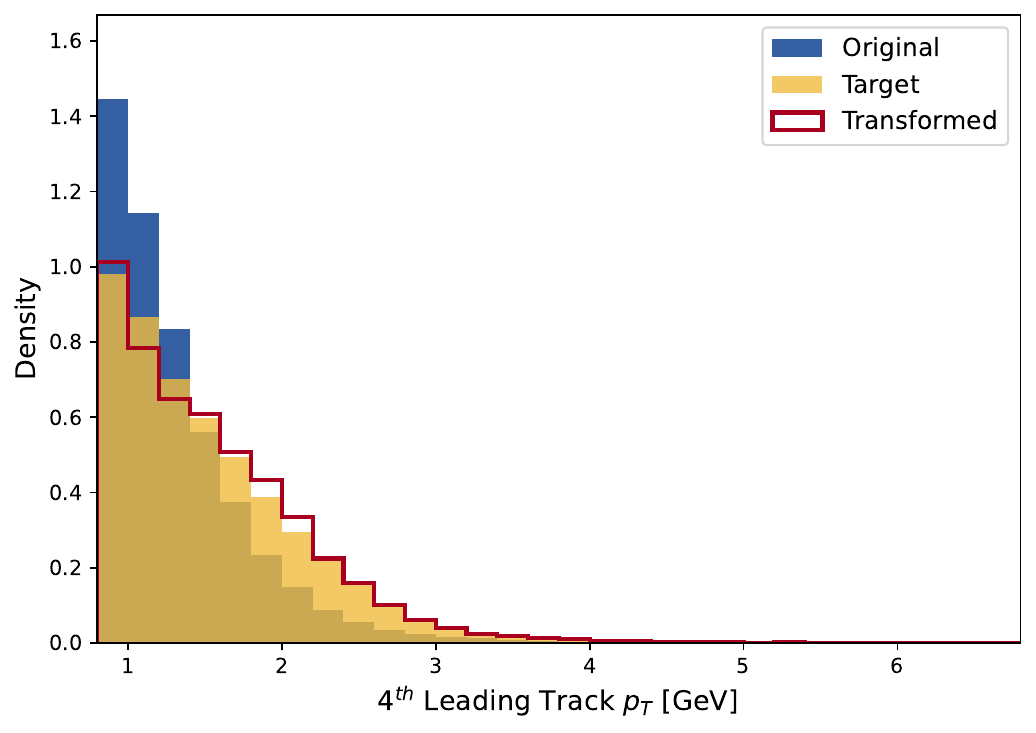}
    \includegraphics[width=0.32\textwidth]{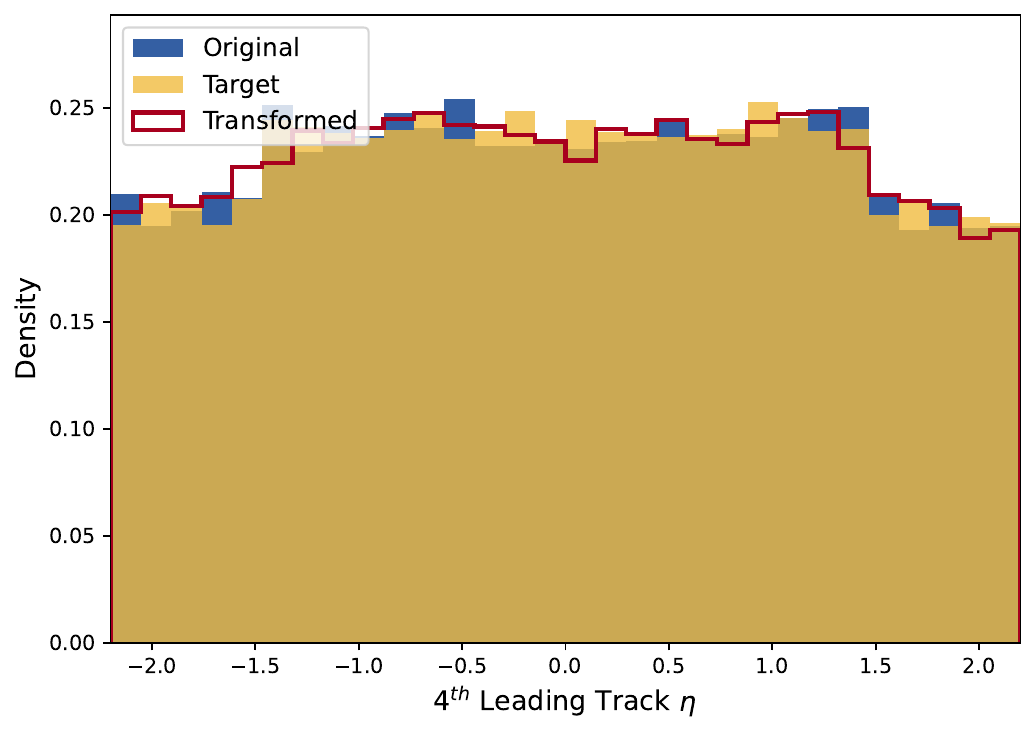}
    \includegraphics[width=0.32\textwidth]{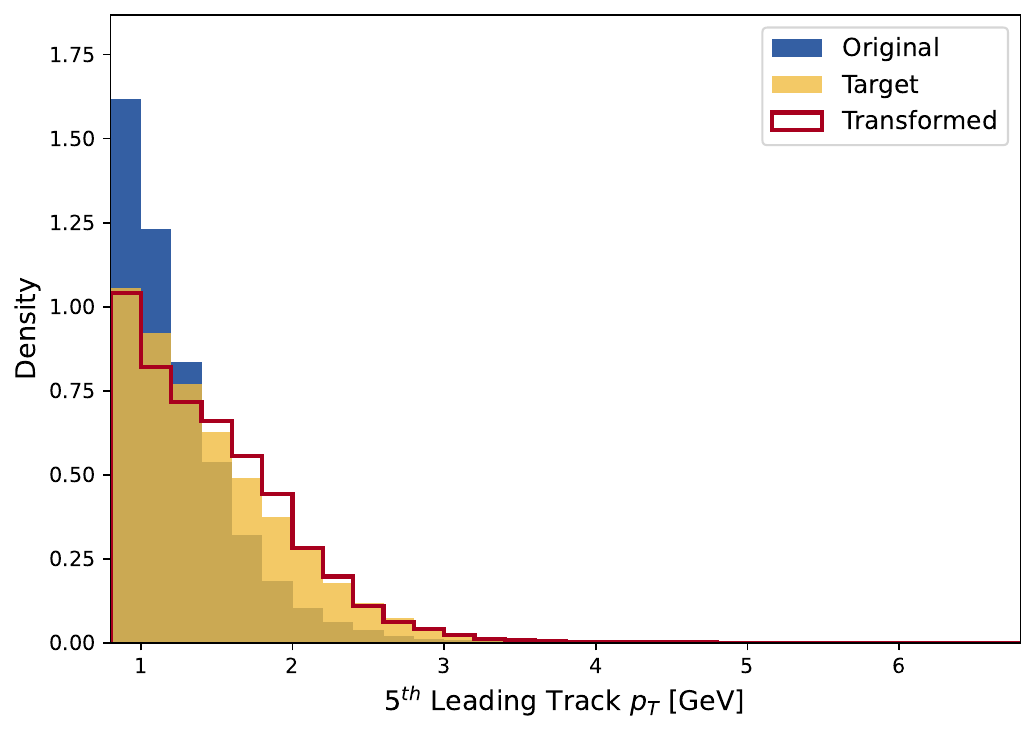}
    \caption{Comparison of selected input features for the underlying event models of the original data-set (blue), the target data-set (yellow) and the transformed results, predicted by the neural network (violet)}
    \label{fig:UEFeaturesTrack}
\end{figure}

\begin{figure}
    \centering
    \includegraphics[width=0.32\textwidth]{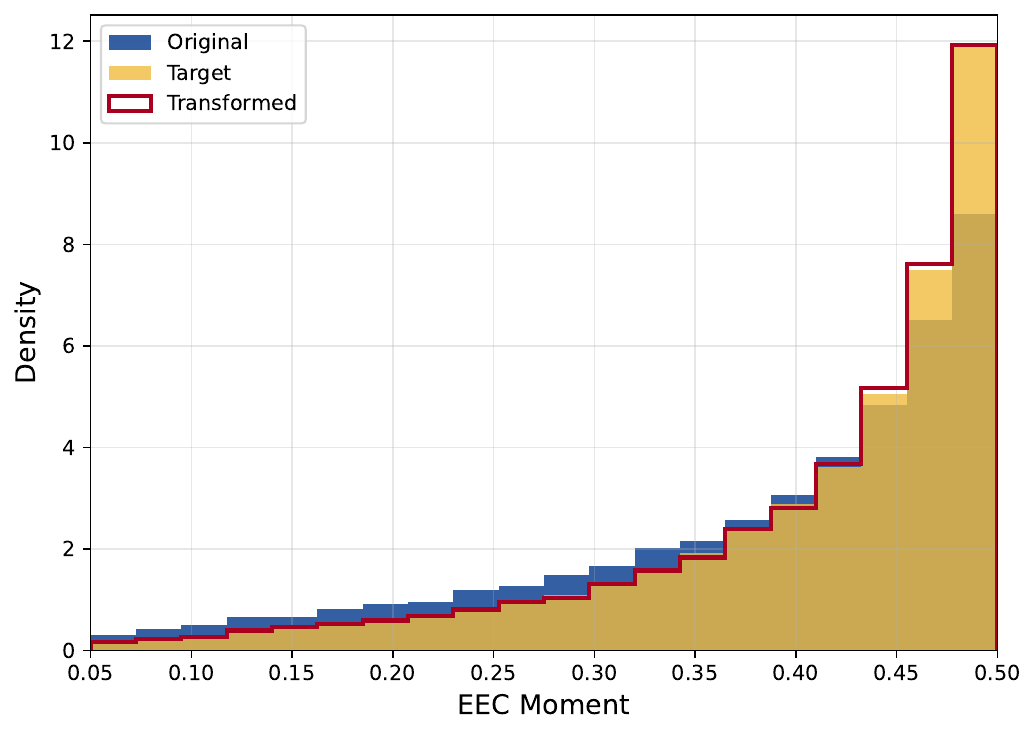}
    \includegraphics[width=0.32\textwidth]{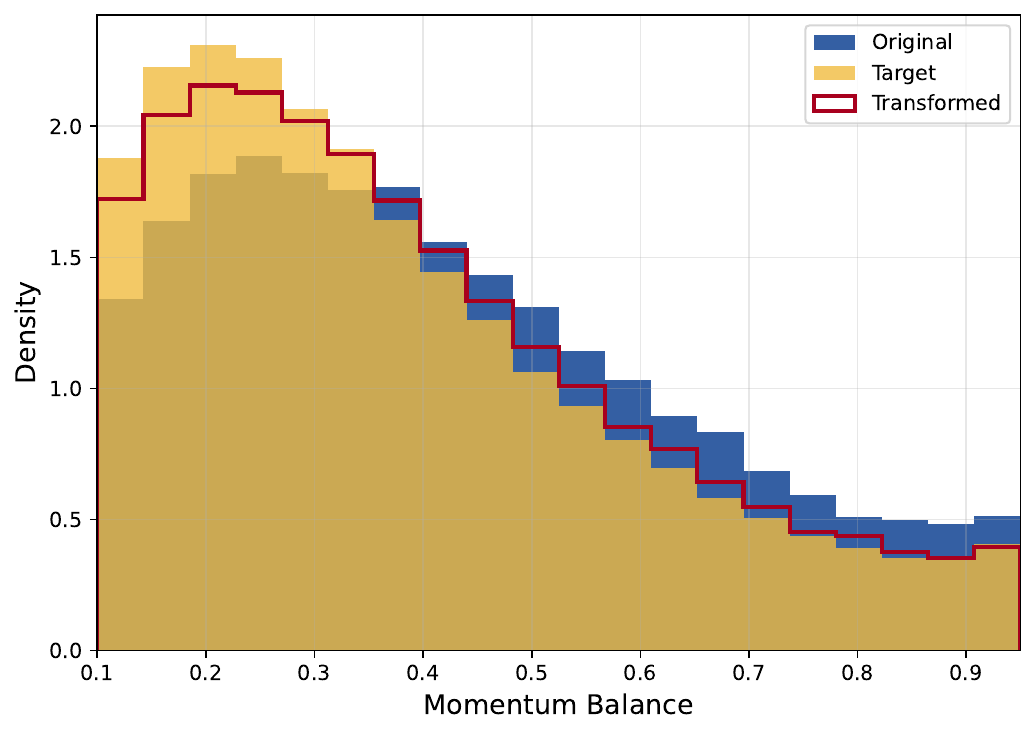}
    \includegraphics[width=0.32\textwidth]{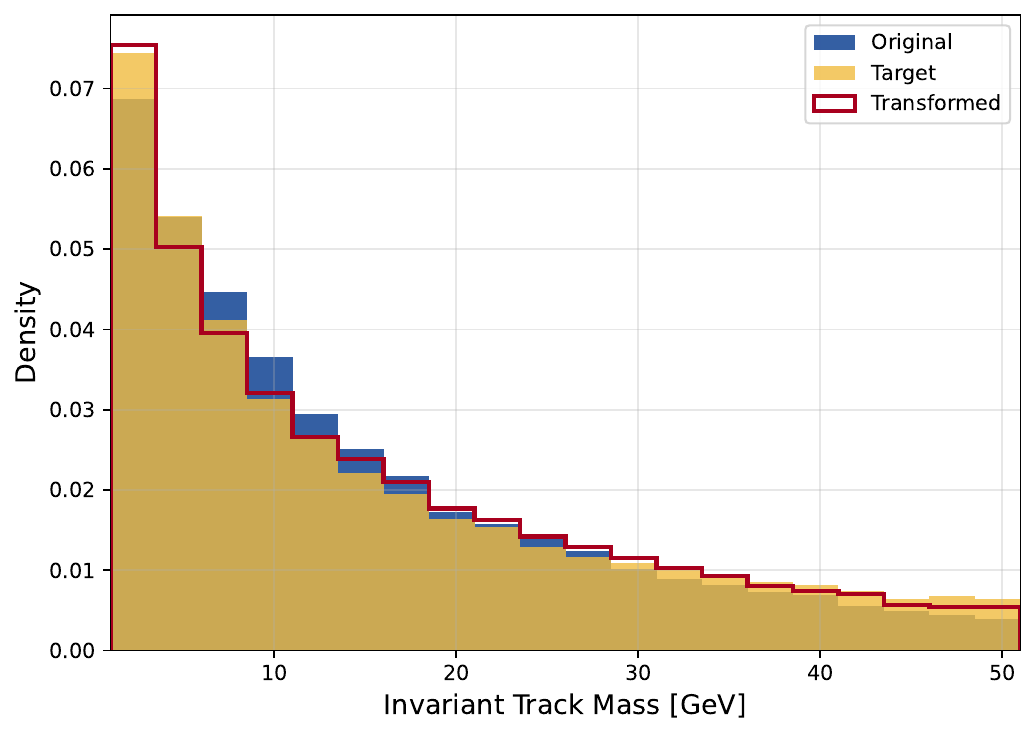}
    \caption{Comparison of derived observables based in the input features for the underlying event models of the original data-set (blue), the target data-set (yellow) and the transformed results, predicted by the neural network (violet)}
    \label{fig:UEDerivedQuantitiesTracks}
\end{figure}

In this scenario, the global residual transformation approach was found to be insufficient, leading to noticeable discrepancies in both individual feature distributions and derived observables. This reflects the increased complexity of the underlying event, where soft and highly correlated structures dominate. Consequently, we focus exclusively on the results obtained with the two-step residual transformation.
In the first stage, feature-wise residual networks are trained to match the one-dimensional distributions of the individual track-level observables. In the second stage, a global refinement network enforces consistency of derived quantities and restores correlations across tracks.
The resulting feature distributions after the full two-step training are shown in Figure~\ref{fig:UEFeaturesTrack}. A very good level of agreement between the transformed and target samples is observed across all displayed features, demonstrating that the method successfully captures the differences between the two tunes at the level of individual track kinematics.
The same conclusion holds for the derived observables, shown in Figure~\ref{fig:UEDerivedQuantitiesTracks}, where excellent agreement is achieved across all considered quantities. This indicates that the second-stage refinement effectively reconstructs higher-level structures from the feature-wise corrections.

\begin{figure}
    \centering
    \includegraphics[width=0.32\textwidth]{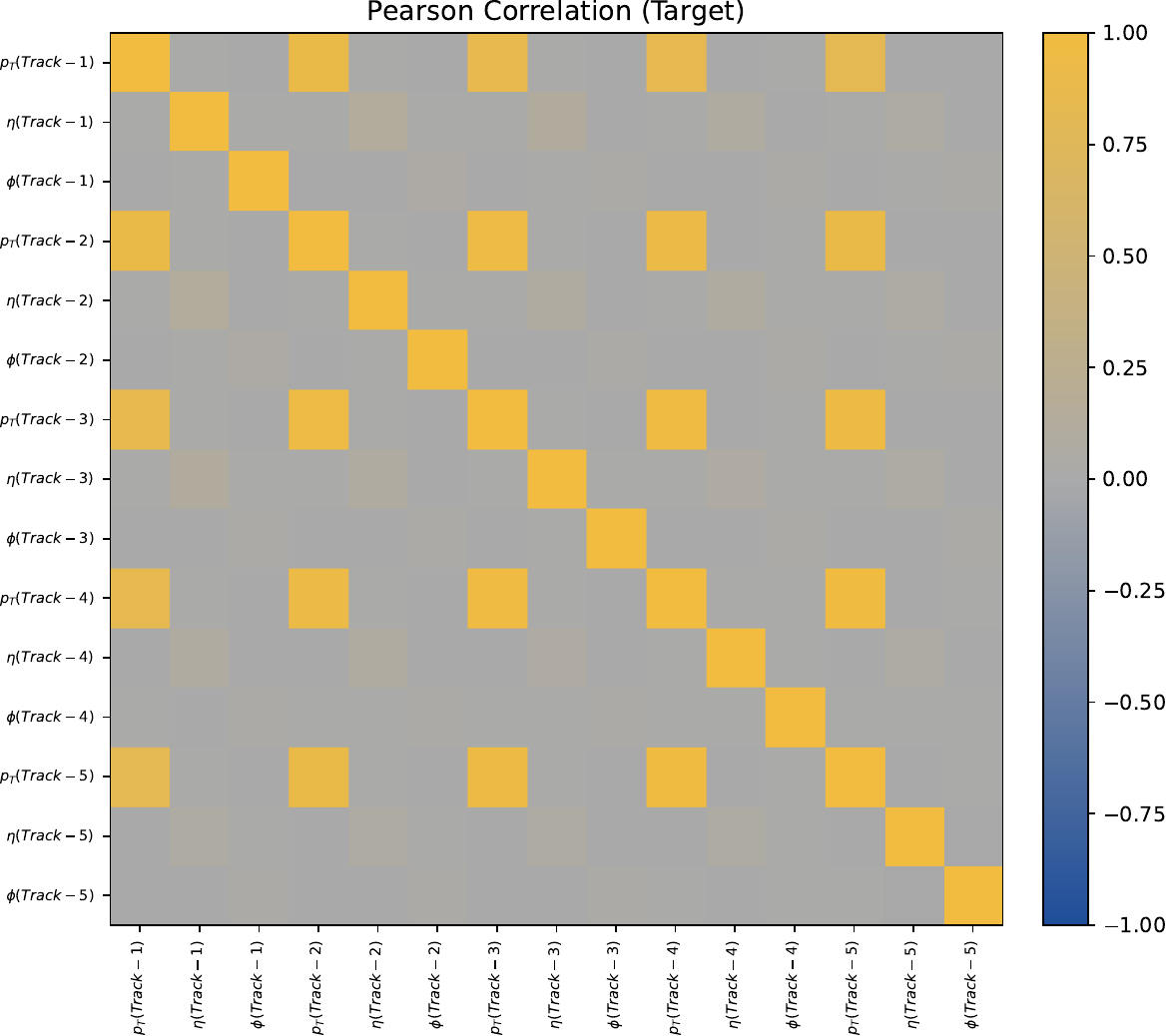}
    \includegraphics[width=0.32\textwidth]{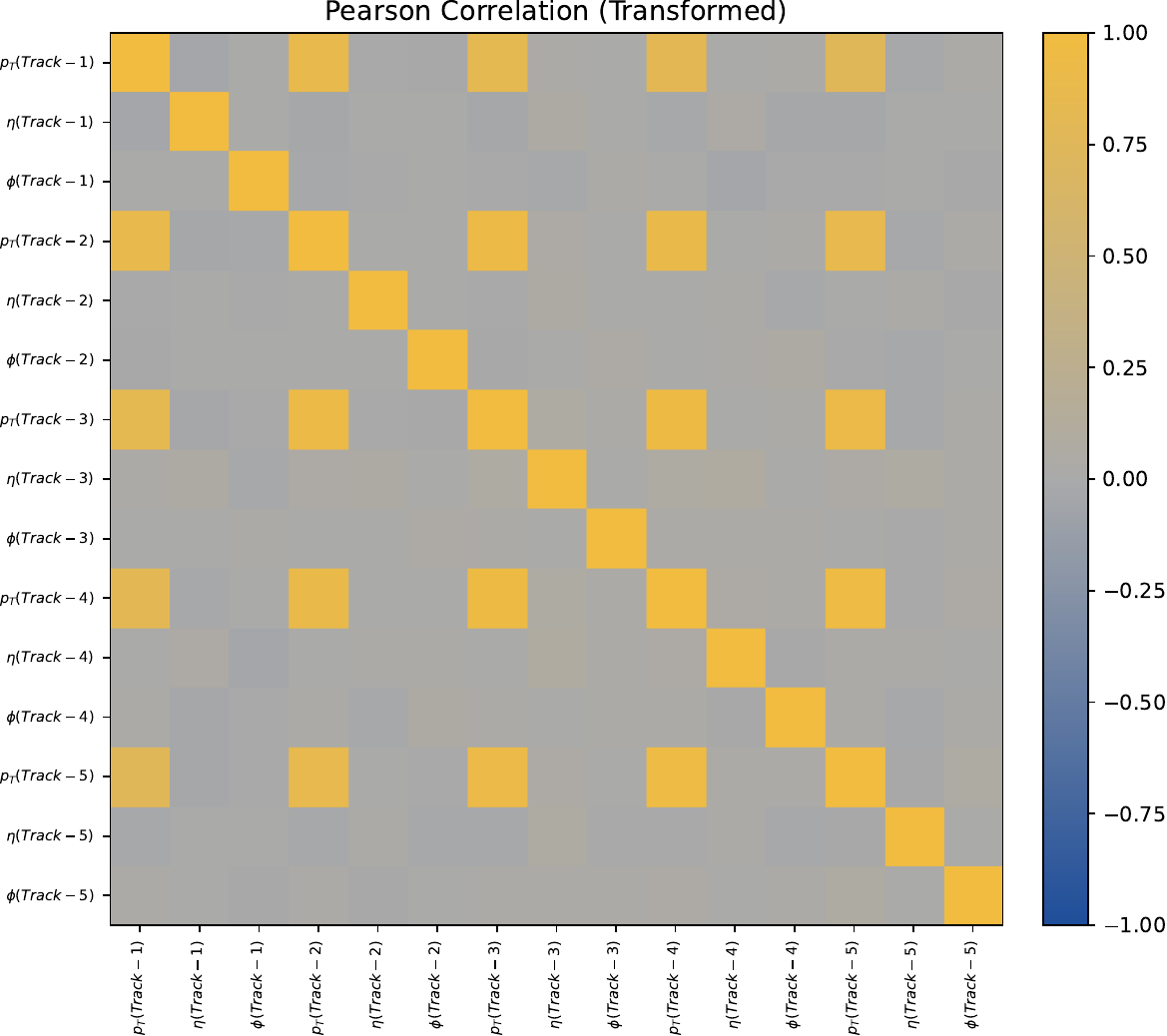}
    \includegraphics[width=0.32\textwidth]{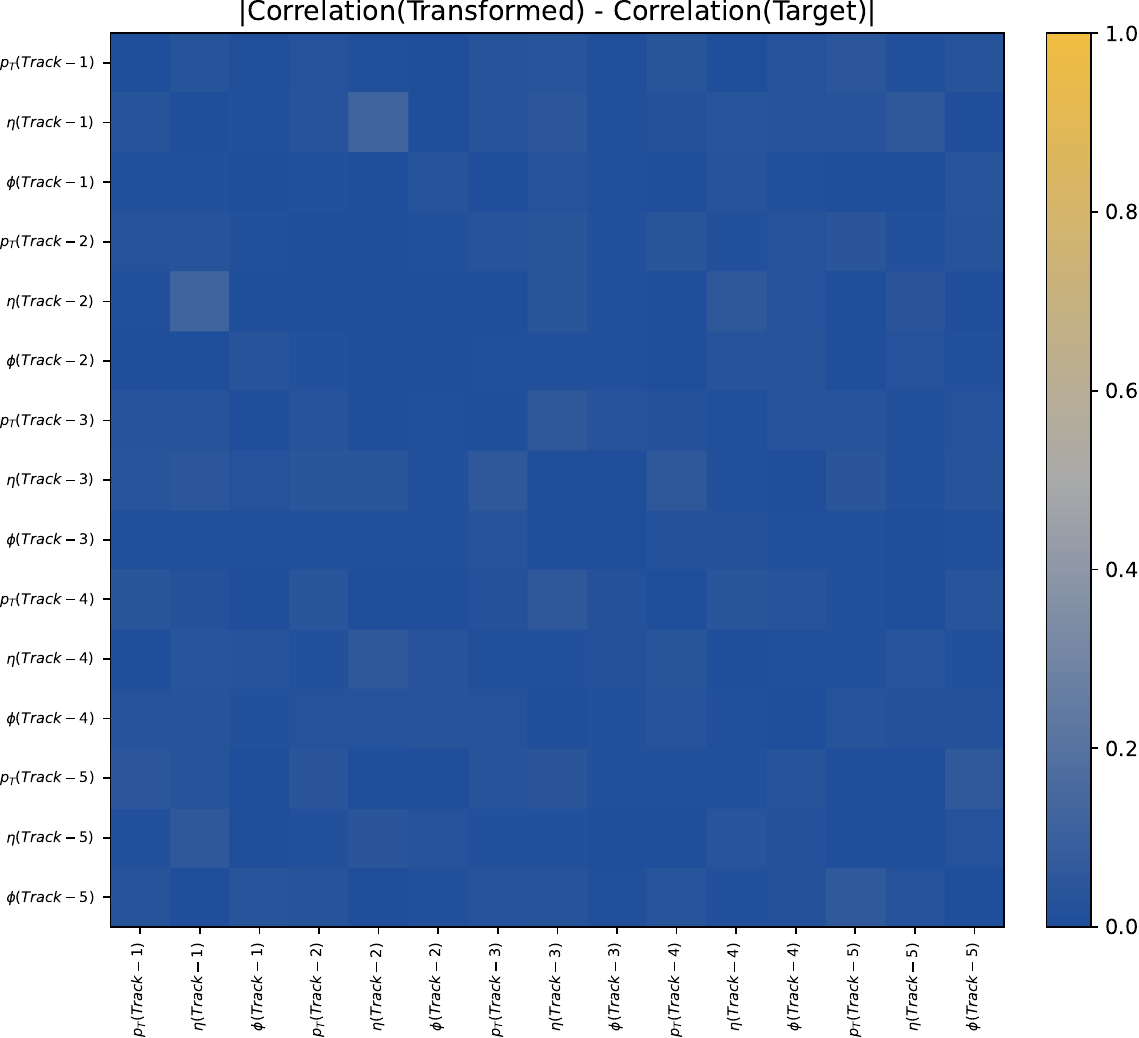}
    \caption{Comparison of Pearson correlation coefficients of the target data-set (left), the transformed results by the neural network (middel) and their difference (right) for the underlying event models.}
    \label{fig:UECorrelationTracks}
\end{figure}

Finally, the correlation structure is examined in Figure~\ref{fig:UECorrelationTracks}, where the Pearson correlation matrices for the first 15 tracks are compared. These tracks dominate the event activity and therefore provide the most relevant contribution to the multivariate structure. The correlations are well preserved, with typical deviations below 0.1. This demonstrates that the two-step approach is capable of maintaining the complex correlation patterns characteristic of underlying event dynamics while accurately reproducing both feature-level and derived distributions.

\subsection{Implications for Classification Problems and Higher Dimensional Correlations}

\begin{figure}
    \centering
    \includegraphics[width=0.49\textwidth]{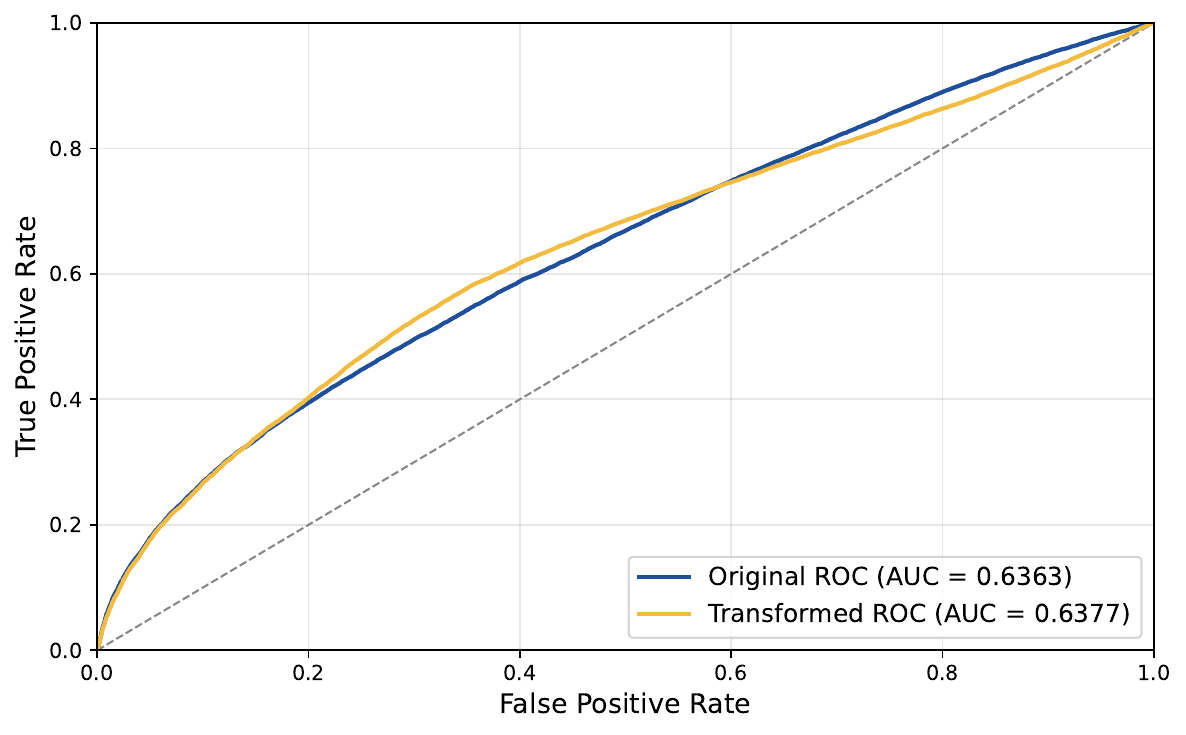}
    \includegraphics[width=0.49\textwidth]{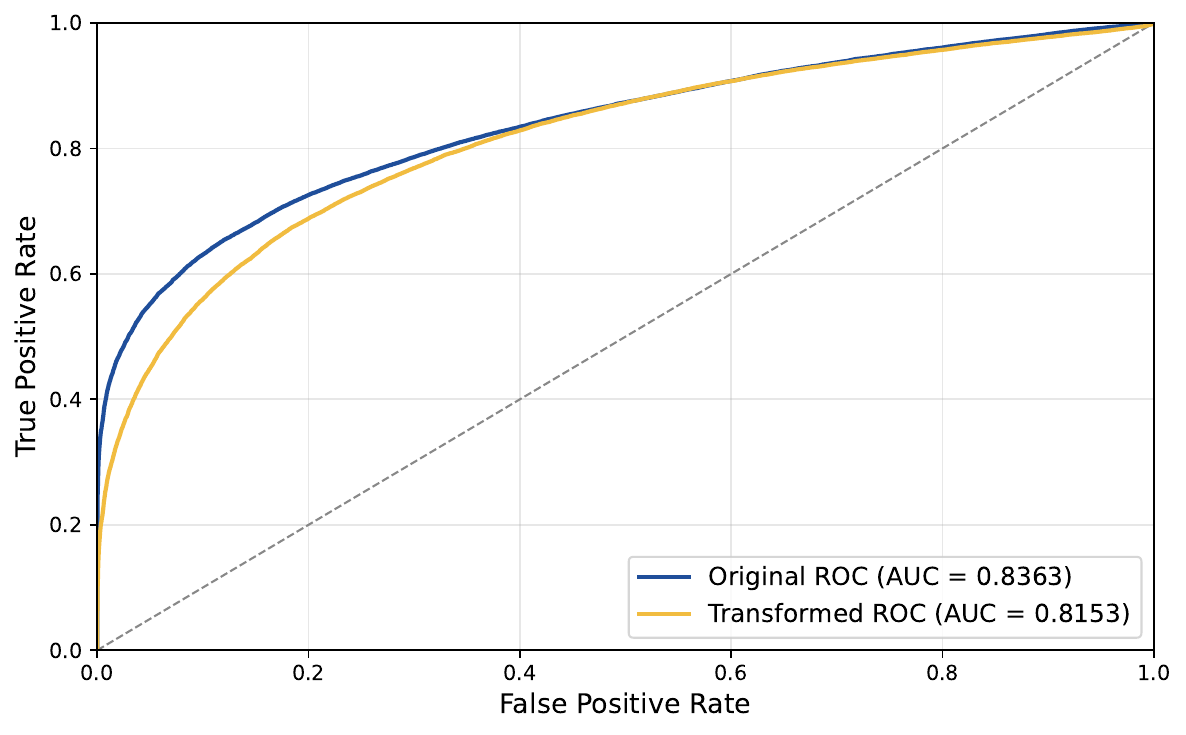}
    \caption{ROC curves for the classification studies using the underlying-event samples (left) and the top-quark pair samples (right). In each panel, the two curves compare the performance of a classifier trained to distinguish the original from the target sample with its corresponding performance when the transformed sample is used in place of the target sample. The similarity of the ROC curves therefore provides a test of how well the transformation reproduces the multivariate structure of the target distribution relevant to the classifier.}
    \label{fig:UEROC}
\end{figure}

So far, the evaluation has focused primarily on one-dimensional distributions and selected derived observables. However, realistic analyses rely on higher-dimensional correlations and often employ multivariate classifiers. It is therefore essential to assess whether the transformed samples also provide an accurate description in such contexts.
As a test case, we train a deep neural network (DNN) classifier to distinguish between the original and the target data sets. The classifier is trained until convergence, with early stopping applied based on the validation performance. After training, the classifier is evaluated on the transformed data set, treating it as a proxy for the target sample.
The resulting receiver operating characteristic (ROC) curves are shown in Figure~\ref{fig:UEROC}. The area under the curve (AUC) obtained when using the transformed sample agrees with that of the true target sample at the level of approximately 1\%. In addition, the full ROC curves exhibit very similar behavior across the entire range.

These results indicate that the transformed data set successfully reproduces not only one-dimensional distributions, but also the higher-dimensional structures relevant for classification tasks. The observed differences are smaller than typical systematic uncertainties in such analyses, suggesting that the proposed approach is suitable for use in realistic multivariate workflows.

\section{Conclusion}

In this work, we introduced a residual learning framework for transforming simulated high-energy physics event samples such that they better reproduce target distributions, while preserving the essential structure of the original data. The approach is motivated by the common situation in which baseline simulations already capture the dominant physics, but require targeted corrections to improve agreement with data or alternative modeling assumptions.
Two complementary strategies were investigated: a global residual transformation, in which a single neural network performs a joint correction of all features, and a two-step residual approach, which separates the learning of one-dimensional feature distributions from the restoration of higher-level correlations and derived observables. Both methods rely on bounded residual updates and differentiable histogram-based losses, enabling stable training and direct optimization of distributional agreement.
The performance of the framework was evaluated in two distinct physics scenarios. In the first case, $t\bar{t}$ production was transformed from Tevatron to LHC conditions, representing a significant shift in kinematic regime. In the second case, differences between underlying event tunes were studied, probing a highly complex and correlation-dominated environment. In both setups, the methods achieved good agreement with the target distributions at the level of individual features and derived observables.
The comparison between the two approaches highlights the advantages of the two-step strategy. While the global model already captures the dominant effects, the two-step approach consistently improves the agreement in one-dimensional distributions and leads to a more accurate preservation of correlations. This is particularly important in complex scenarios such as underlying event modeling, where the global approach alone was found to be insufficient.
Beyond one-dimensional comparisons, the quality of the transformed samples was further validated using a classification-based test. A neural network trained to distinguish between original and target samples shows nearly identical performance when evaluated on the transformed data, with differences in the area under the ROC curve at the level of $\mathcal{O}(1\%)$. This demonstrates that the proposed method successfully captures higher-dimensional structures relevant for realistic multivariate analyses.

It should be noted that the studies presented here are based on comparatively small data samples. In realistic experimental environments, significantly larger simulated data sets are available. Access to such large-scale data could enable the use of more expressive architectures, such as transformer-based models, which may be able to learn richer internal representations of the underlying physics. In particular, such models could improve the ability to extrapolate from limited information about target distributions and more effectively capture complex, high-dimensional dependencies.
Overall, the results indicate that residual transformation models provide a flexible and robust tool for improving simulated event samples. The framework is particularly well suited for scenarios in which only limited information about the target distribution is available, such as cases where primarily one-dimensional measurements are known, while correlations remain only partially constrained.

Future work could explore extensions to more complex architectures, alternative loss formulations, and applications to real data, including uncertainty estimation and systematic variations. In this context, the proposed approach may offer a promising new avenue for refining simulations and improving the modeling of complex final states in high-energy physics.

\bibliographystyle{unsrt}
\bibliography{./Bibliography}

\end{document}